\documentclass[journal]{IEEEtran}

\usepackage{cite}
\usepackage{balance}   
\usepackage[colorlinks=true,
            linkcolor=blue,
            citecolor=blue,
            urlcolor=blue]{hyperref}
\usepackage{url}
\usepackage{amsmath,amssymb,amsfonts}
\usepackage{bbm}     
\usepackage{graphicx}
\usepackage{fancyhdr}
\usepackage{cases}
\usepackage{textcomp}
\usepackage{extarrows}
\usepackage{orcidlink}

\usepackage{multirow}
\usepackage{caption}
\usepackage{float}
\usepackage{subcaption}
\usepackage{makecell}

\usepackage{booktabs}
\usepackage{array}
\usepackage{footnote}
\makesavenoteenv{tabular}

\usepackage{algorithm,algpseudocode}

\usepackage{multirow,tabularx}
\usepackage{multicol,float,subcaption}
\usepackage{mathtools}
\usepackage{xcolor}
\usepackage[english]{babel}
\addto\captionsenglish{\renewcommand{\figurename}{Fig.}}
\usepackage{amsthm}

\usepackage{setspace}

\makeatletter
\renewcommand{\maketag@@@}[1]{\hbox{\m@th\normalsize\normalfont#1}}%
\makeatother

\def\BibTeX{{\rm B\kern-.05em{\sc i\kern-.025em b}\kern-.08em
    T\kern-.1667em\lower.7ex\hbox{E}\kern-.125emX}}

\begin{document}

\title{TelecomGPT-R1: Unified Post-Training for Reasoning Across Heterogeneous Telecom Tasks}

\author{Bohao Wang,
Chenwei Wu,
Hang Zou,
Yu Tian,
Lina Bariah,
Li Wei,
Chongwen Huang,
Yongliang Shen,
Zhaoyang Zhang,
and M\'{e}rouane Debbah,~\IEEEmembership{Fellow,~IEEE}

\thanks{\textit{Corresponding author: Yongliang Shen, Hang Zou, Chongwen Huang.}}
\thanks{B. Wang, L. Wei, C. Huang, and Z. Zhang are with the College of Information Science and Electronic Engineering, Zhejiang University, 310027 Hangzhou, China (e-mail: \{\href{mailto:bohaowang@zju.edu.cn}{bohaowang}, 
\href{mailto:l_wei@zju.edu.cn}{l\_wei}, \href{mailto:chongwenhuang@zju.edu.cn}{chongwenhuang}, \href{mailto:ning_ming@zju.edu.cn}{ning\_ming}\}@zju.edu.cn).}
\thanks{C. Wu is with the Department of Electrical Engineering and Computer Science, University of Michigan, Ann Arbor, MI 48109-2122, USA (e-mail: \href{mailto:chenweiw@umich.edu}{chenweiw@umich.edu}).}
\thanks{B. Wang, H. Zou, Y. Tian, L. Bariah, and M. Debbah are with the Research Institute for Digital Future, Khalifa University, P O Box 127788, Abu Dhabi, UAE (e-mail: \{\href{mailto:bohao.wang@ku.ac.ae}{bohao.wang}, \href{mailto:hang.zou@ku.ac.ae}{hang.zou}, \href{mailto:yu.tian@ku.ac.ae}{yu.tian}, \href{mailto:lina.bariah@ku.ac.ae}{lina.bariah}, \href{mailto:merouane.debbah@ku.ac.ae}{merouane.debbah}\}@ku.ac.ae).}
\thanks{Y. Shen is with the College of Computer Science and Technology, Zhejiang University, 310027 Hangzhou, China (e-mail: \href{mailto:syl@zju.edu.cn}{syl@zju.edu.cn}).}
}
\maketitle

\begin{abstract}
Large language models (LLMs) offer great potential to automate a broad range of telecom engineering tasks by reasoning over standards, network configurations, mathematical models, source code, and
operational logs. However, existing telecom LLMs still struggle to reliably reason across these diverse tasks and data types.  General-purpose LLMs often lack
reliable grounding in telecom-specific knowledge, while
telecom-specialized models are typically developed for narrower task families and exhibit limited multi-task performance. To fill this gap, we introduce \textit{TelecomGPT-R1}, a family of open source unified telecom reasoning models structured around four complementary axes: protocol, knowledge, modeling, and fault. 
We first develop an axis-aware data generation framework that refines coarse public telecom artifacts into verified question--answer pairs and high quality chain-of-thought (CoT) reasoning trajectories, yielding a training corpus containing 104{,}880 examples. Building on this corpus, supervised fine-tuning (SFT) instills telecom knowledge and evidence-grounded reasoning patterns to overcome the cold start barrier for reinforcement learning (RL). We then apply dynamic sampling policy optimization (DAPO) with task-routed rubric rewards to keep RL updates informative and stable across heterogeneous telecom reasoning tasks. These rewards decompose axis-specific CoT traces into verifiable reasoning units and combine grounded dense process credit with outcome correctness, allowing RL to learn generalizable problem solving behaviors from verifiable telecom evidence. We release the \textit{TelecomGPT-R1} models and a reproducible training recipe to support further community development. 
Evaluations on seven benchmarks of the GSMA Open Telco Leaderboard show that the open-source \textit{TelecomGPT-R1-27B} achieves an 89.64\% mean score, outperforming leading proprietary models, including GPT-5, Claude, and Gemini.

\end{abstract}


\section{Introduction}\label{sec:intro}

Recent advances in post-training methods such as reinforcement learning with verifiable rewards (RLVR)~\cite{openai2024o1,deepseekr1,dapo2025} have transformed large language models (LLMs) into increasingly capable reasoners, enabling complex problem solving on general domain tasks, like coding and mathematical reasoning~\cite{openai2024o1,deepseekr1,dapo2025}. This progress has promising potential for the telecom domain, where a capable reasoning LLM can support engineers in tasks that currently require deep expert knowledge, including engineering tasks that today demand scarce human expert resources, such as standards interpretation, protocol analysis, log diagnosis, configuration inspection, wireless formula verification, and code-level troubleshooting \cite{bariah2024commag,zou2026nree}.

However, current use of LLMs in telecom is limited, due to the fact that existing LLMs perform well on static question-answer (QA) tasks, but real world telecom engineering demands reliable reasoning over standards, logs, configurations, protocol traces, equations, and codes. 
For example, diagnosing a network failure may require jointly interpreting a 3GPP procedure and an O-RAN configuration table~\cite{gajjar2025oransight}, tracing log anomalies, checking key performance indicator (KPI) shifts against expected protocol behavior, and verifying that the implementation and configuration comply with relevant standards.
Accordingly, these workflows expose a new challenge related to the heterogeneity of telecom data formats and representations. Hence, an effective telecom LLM must be capable of \textit{unified reasoning} over diverse tasks and data sources, where each is characterized by different evidence structures, reasoning procedures, and correctness criteria.

Neither existing general-purpose models nor telecom domain LLMs meet this requirement. On one hand, frontier reasoning models such as GPT-5.4~\cite{openai2025gpt5} and Claude Opus 4.6~\cite{anthropic2025claude45} demonstrate strong general domain reasoning, and mathematical and coding abilities. However, they lack domain knowledge and are not trained to reliably ground their reasoning in telecom-specific standards, counters, protocol procedures, or operational inputs. As shown in Section~\ref{sec:analysis:coldstart}, such models can produce fluent and internally consistent explanations, while relying on incorrect assumptions about 3GPP protocols and hallucinated fault evidence. These failures indicate that general reasoning capability alone is insufficient for telecom tasks, which require detailed domain knowledge and strict adherence to standards-defined procedures and constraints.

On the other hand, telecom-specialized models improve
domain coverage, but most existing efforts are focused on developing and evaluating Telecom LLMs for a narrow set of tasks.
TelecomGPT~\cite{telecomgpt2024}, Tele-LLMs~\cite{maatouk2026tele}, and TelcoLM~\cite{barboule2024telcolm} adapt general-purpose LLMs through continued pre-training,
instruction tuning, or alignment on telecom corpora, in order to
strengthen telecom knowledge understanding and general-domain
question answering. Standards-oriented systems employ retrieval, long-context support, or instruction tuning for grounded answering over 3GPP and O-RAN documents~\cite{bornea2024telcorag,yilma2024telecomrag,nikbakht2024tspecllm,erak2024teleoracle,bornea2025telcoorag,gajjar2025oransight}. Other efforts specialize in network prediction and optimization, wireless mathematical reasoning, or mobile-network fault diagnosis~\cite{wu2024netllm,wirelessmathlm2025,sana2025reasoning}. Although these methods improve performance on individual tasks, is it not yet clear if a single model can learn the different reasoning procedures required across telecom standards, logs, configurations, equations, and code. In Section~\ref{sec:analysis:multisource}, we show that training on a single telecom task does not transfer reliably to other tasks and can negatively impact other telecom capabilities, demonstrated in both task performance and reasoning behavior. 
These results motivate the joint training across multiple telecom tasks and data sources. However, such training introduces new challenge as the tasks differ substantially in their chain-of-thought (CoT) structures, reasoning lengths, verifier designs, and reward density~\cite{teh2017distral,schaul2019ray,hessel2019multitask}. Therefore, the key challenge is to develop a single telecom-specialized model with broad, reliable, and verifiable reasoning capabilities across heterogeneous telecom tasks and data sources.

\begin{figure}[t]
    \centering
    \includegraphics[width=0.93\columnwidth]{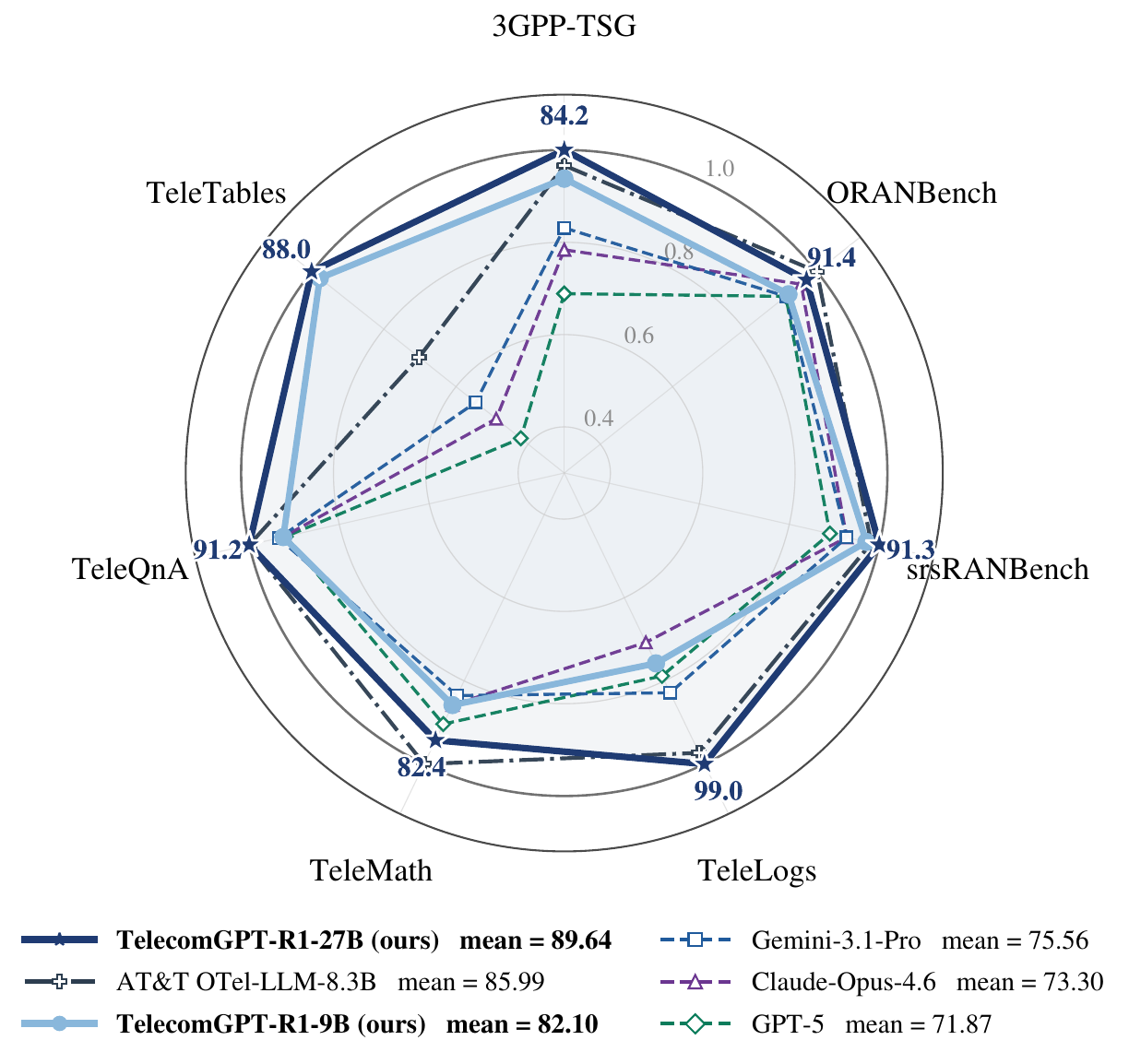}
    \caption{\textit{TelecomGPT-R1} family on the GSMA Open Telco Leaderboard. The 27B model ranks \#1 overall with an $89.6\%$ mean score.}
    \label{fig:radar}
\end{figure}

To overcome this challenge, we introduce \textit{TelecomGPT-R1}, a family of open source unified telecom reasoning models post-trained from Qwen3.5~\cite{qwen35}. Instead of being optimized for a single telecom task, \textit{TelecomGPT-R1} models are trained as a unified policy over heterogeneous telecom technical artifacts, including standards, logs, configuration tables, KPIs, formulas, and code. 
We organize telecom reasoning into four axes: \textit{protocol}, \textit{knowledge}, \textit{modeling}, and \textit{fault}. These axes cover procedure analysis grounded in standards, disambiguation of telecom facts and concepts, mathematical derivations together with reasoning over code and tables, and operational fault diagnosis. The key principle is to keep the model unified while adapting data construction, verification, and reward design to the requirements of each axis.

Following this principle, we construct a 104{,}880-instance telecom reasoning corpus, comprising 92{,}380 supervised fine-tuning (SFT) examples and 12{,}500 reinforcement learning (RL) rollout prompts. Using an axis-aware generation and refinement pipeline, we convert heterogeneous public telecom source materials, including standards, logs, configuration tables, formulas, and operational records, into verified QA pairs. We further collect high quality CoT reasoning trajectories through telecom knowledge-augmented distillation of multiple teacher models and robust quality filtering.

Building on this corpus, we adopt a two-stage post-training recipe. First, SFT equips the base model with telecom knowledge and task-specific reasoning patterns, providing a suitable initialization for RL. Second, RL moves the SFT model beyond memorization of telecom knowledge and imitation of distilled CoT traces, enabling it to learn generalizable problem solving behaviors from verifier-grounded feedback. Specifically, we apply dynamic sampling policy optimization (DAPO) \cite{dapo2025} with task-routed rubric rewards, which combine dense process-level feedback with outcome correctness. This enables RL to improve telecom reasoning while constraining policy updates using verifiable task signals.

Through systematic studies of data construction, multi-source training, reward design, and scaling, we identify four principles for building strongly capable telecom reasoners. First, telecom reasoning faces a pronounced cold start problem. Base models and even strong teachers can produce fluent but factually incorrect traces, making knowledge augmented and failure mined SFT necessary before RL can provide useful learning signals. Second, broad competence requires diverse source training. Single source training produces localized gains but weak or negative transfer, whereas multi-source training enables one policy to learn complementary reasoning modes. Third, difficult tasks require verifier grounded dense rewards. Outcome only rewards are too sparse for protocol and fault reasoning, while unconstrained dense rewards can reinforce hallucinated rules, metrics, or standards claims. Finally, scaling sample size and model capacity consistently improves downstream task performance.

The main contributions are summarized below.
\begin{itemize}
    \item We design a unified framework for heterogeneous telecom reasoning. Specifically, we introduce \textit{TelecomGPT-R1} and an axis-aware framework that combines verified multi-source data construction, multi-teacher SFT, and verifier-grounded RL with task-routed dense rubric rewards to train a single policy across protocol, knowledge, modeling, and fault reasoning.

    \item We establish empirically validated principles for training telecom reasoners. Through controlled experiments, we show that broad telecom reasoning requires a knowledge-grounded cold start, source-diverse training, verifier-grounded process rewards for difficult tasks, and complementary scaling of model capacity and data coverage.

    \item We construct a verified multi-source telecom reasoning corpus from coarse public artifacts. Through source-matched question construction, CoT generation, and verification, we produce 104{,}880 examples spanning the four reasoning axes and heterogeneous telecom sources.

    \item We achieve state-of-the-art performance and release open resources. Across seven benchmarks on the GSMA Open Telco Leaderboard, \textit{TelecomGPT-R1-27B} achieves an $89.64\%$ mean score, outperforming leading proprietary models, including GPT-5, Claude, and Gemini. We release the models, code, training recipe, and evaluation tools to support reproducible research.
\end{itemize}

\section{Problem Formulation}
\label{sec:problem}

\textit{TelecomGPT-R1} formulates reasoning across heterogeneous telecom tasks as learning one autoregressive policy under axis-specific data, reasoning, verification, and reward structures. The problem has two coupled components, i) constructing aligned SFT and RL datasets from heterogeneous telecom sources, and ii) developing a unified post-training strategy that first establishes a reliable supervised initialization and then further optimizes the shared policy using task-dependent verifier feedback.

Accordingly, we define four reasoning axes:
\begin{equation}
\label{eq:axis_set}
\mathcal{A}
=
\{
\mathrm{protocol},
\mathrm{knowledge},
\mathrm{modeling},
\mathrm{fault}
\}.
\end{equation}
Each axis $a\in\mathcal{A}$ has distinct data, reasoning, verification, and reward structures, routed by its axis tag under a shared policy.

Let $\mathcal{S}$ denote the collection of public telecom sources. The data construction problem is expressed as
\begin{equation}
\label{eq:data_construction}
\left(
\mathcal{D}_{\mathrm{SFT}},
\mathcal{D}_{\mathrm{RL}}
\right)
=
\mathcal{G}
\left(
\mathcal{S};
\mathcal{A},
\{Q^{(a)}\}_{a\in\mathcal{A}}
\right),
\end{equation}
where $\mathcal{G}$ is the axis-aware construction procedure and $Q^{(a)}$ determines
whether a candidate response is accepted for axis $a$. It produces
\begin{equation}
\label{eq:training_datasets}
\begin{aligned}
\mathcal{D}_{\mathrm{SFT}}
&=
\left\{
(x_n,o_n^\star,y_n^\star,a_n,m_n)
\right\}_{n=1}^{N_{\mathrm{SFT}}},\\
\mathcal{D}_{\mathrm{RL}}
&=
\left\{
(x_n,y_n^\star,a_n,m_n)
\right\}_{n=1}^{N_{\mathrm{RL}}},
\end{aligned}
\end{equation}
where $n$ indexes an item, $N_{\mathrm{SFT}}$ and $N_{\mathrm{RL}}$ are the dataset sizes of SFT and RL respectively, $x_n$ is the task input, $o_n^\star$ is an accepted reasoning trace, $y_n^\star$ is the verifier-readable target, $a_n\in\mathcal{A}$ is the axis tag, and $m_n$ contains verifier metadata.

For a candidate trace $\widetilde{o}$, the acceptance gate is
\begin{equation}
\label{eq:acceptance_criterion}
Q^{(a)}(x,\widetilde{o};y^\star,m)
=
C^{(a)}(x,\widetilde{o},m)\,
V^{(a)}(\widetilde{o};y^\star,m),
\end{equation}
where curation gate $C^{(a)}\in\{0,1\}$ checks structural validity, evidence completeness, and answer leakage, while the final answer verifier $V^{(a)}$ scores axis-specific correctness of an output $o$ as
\begin{equation}
\label{eq:verifier}
V^{(a)}(o;y^\star,m)
=
\begin{cases}
1, & \text{final answer verified correct},\\
0, & \text{otherwise}.
\end{cases}
\end{equation}
If $Q^{(a)}=1$, $\widetilde{s}$ becomes the SFT target $s^\star$. The aligned RL item omits $s_n^\star$ so that the policy must generate
its own response during rollout. Thus, offline curation and online
reward computation use same task targets and verifier metadata.
The accepted traces in $\mathcal{D}_{\mathrm{SFT}}$ provide supervised
targets for fitting the shared policy. Let $\pi_\theta$ denote the
autoregressive policy parameterized by the trainable parameters $\theta$.
SFT obtains the optimized parameters by solving~\cite{deepseekr1}
\begin{equation}
\label{eq:sft_policy}
\theta_{\mathrm{SFT}}
=
\arg\min_{\theta}
\mathcal{L}_{\mathrm{SFT}}
\left(
\theta;\mathcal{D}_{\mathrm{SFT}}
\right).
\end{equation}
Here, $\mathcal{L}_{\mathrm{SFT}}$ is the autoregressive token loss over
each complete accepted response $s^\star$, and $\theta_{\mathrm{SFT}}$ is
the optimized parameter set. By fitting the conditional distribution of
verified reasoning traces, the resulting policy
$\pi_{\theta_{\mathrm{SFT}}}$ captures axis-appropriate knowledge,
reasoning structures, and output formats and serves as the initialization
and reference policy for RL.

For each $(x,y^\star,a,m)\in\mathcal{D}_{\mathrm{RL}}$, the policy generates a response $o$, and the axis tag selects the verifier reward $R^{(a)}(o;y^\star,m)$. The unified policy optimization problem is \cite{deepseekr1}
\begin{equation}
\label{eq:posttraining_objective}
\begin{aligned}
\theta^\star
={}&
\arg\max_{\theta}\;
\mathbb{E}_{(x,y^\star,a,m)\sim\mathcal{D}_{\mathrm{RL}}}
\Bigg[\\
&\quad\mathbb{E}_{o\sim\pi_\theta(\cdot\mid x)}
\left[
R^{(a)}(o;y^\star,m)
\right]\\
&\quad-
\beta
D_{\mathrm{KL}}
\left(
\pi_\theta(\cdot\mid x)
\parallel
\pi_{\theta_{\mathrm{SFT}}}(\cdot\mid x)
\right)
\Bigg],
\end{aligned}
\end{equation}
where $\theta^\star$ denotes the optimized parameters,
$D_{\mathrm{KL}}$ measures deviation from
$\pi_{\theta_{\mathrm{SFT}}}$, and $\beta\geq0$ controls the regularization strength.

\section{\textit{TelecomGPT-R1} Unified Training Policy}
\label{sec:method}
This section details the two-stage recipe of \textit{TelecomGPT-R1}, source-to-corpus construction followed by SFT and DAPO-based post-training.

\subsection{Source-to-Corpus Construction}
\label{sec:data}

\begin{figure*}[t]
    \centering
    \includegraphics[width=0.95\textwidth]{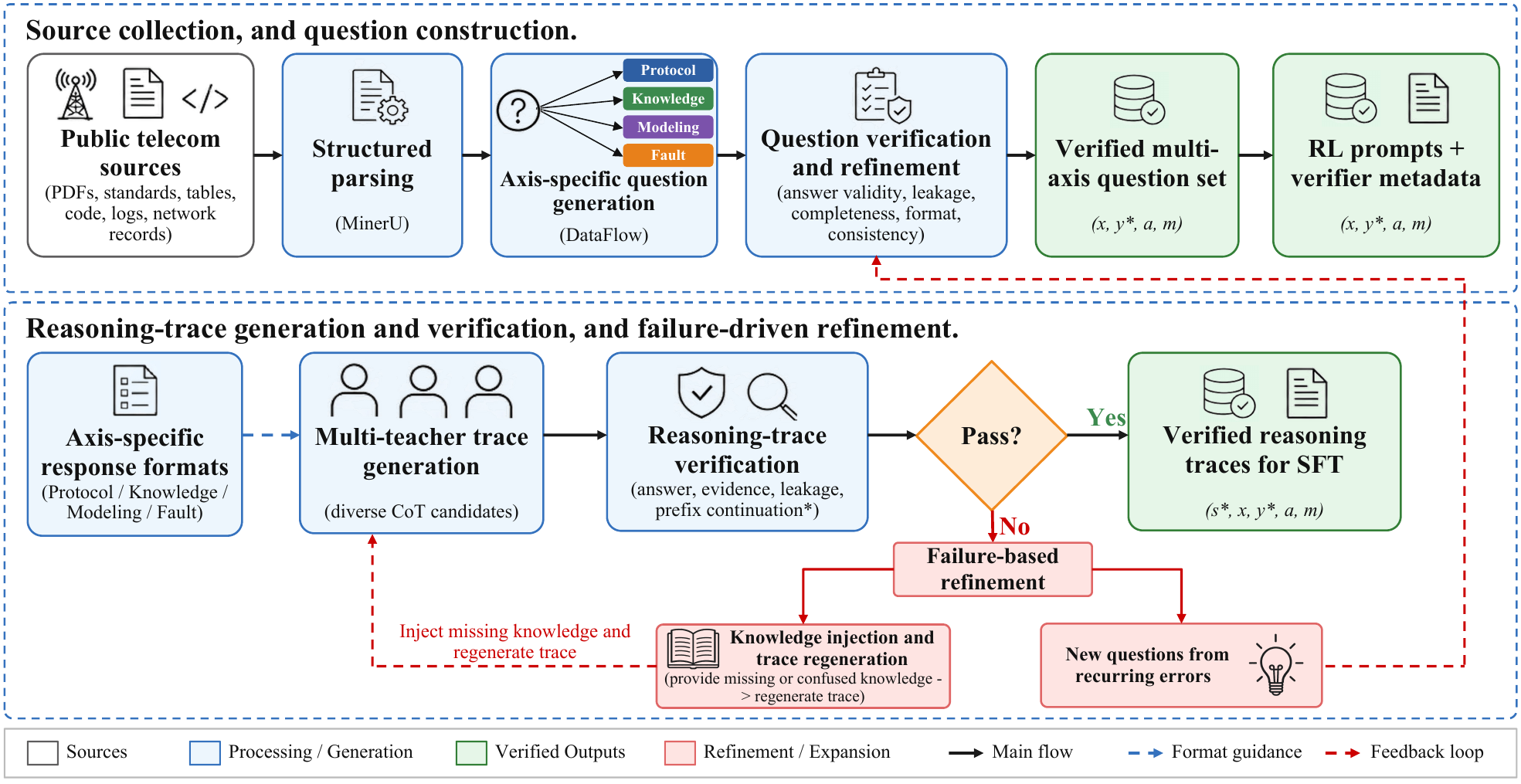}
    \caption{Four steps source-to-corpus construction pipeline for \textit{TelecomGPT-R1}.}
    \label{fig:data_pipeline}
\end{figure*}

Telecom reasoning covers heterogeneous source types, reasoning procedures, and verification methods. Therefore, we construct the corpus through four steps as follows. Fig.~\ref{fig:data_pipeline} summarizes the pipeline.

\subsubsection{Source Collection}

For each reasoning axis, we collect public sources that preserve its native evidence structure. Protocol sources include 3GPP specifications, technical reports, and public O-RAN specifications. Knowledge sources include public multiple-choice question (MCQ) datasets, research papers, operator documentation, and telecom glossaries. Modeling sources include telecom textbooks, open-source calculation datasets, and srsRAN source code. Fault sources pair drive-test measurements with network engineering records.
MinerU~\cite{wang2024mineru} parses these sources into structured units, while DataFlow~\cite{dataflow} supports question generation and format normalization.

\subsubsection{Question Construction}

We convert the collected sources into axis-specific questions as $\mathcal{D}_{\mathrm{SFT}}$ and $\mathcal{D}_{\mathrm{RL}}$.

For the
protocol axis, we extract functions, interfaces, and procedures from
standards and formulate questions identifying their responsible SA, CT, or
RAN working groups (WGs). For the knowledge axis, we standardize public telecom
MCQs and add factual and terminology questions.

For the modeling axis, we construct mathematical, code-grounded, and
table-grounded questions requiring structured derivation. Mathematical items
are first retained with complete questions and answers. Then, items requiring unavailable context are removed, while multi-part exercises are
decomposed into self-contained subproblems with separately verifiable
targets. 
When a source provides formulas or derivations without an explicit
question, we mask a formula, parameter, intermediate result, or final
quantity and ask the model to recover it from the remaining context.
Code-grounded items derive function, call-graph, parameter, and
protocol-mapping questions from program structure, abstract symbol trees, Doxygen documentation, and
protocol-linked symbols. Table-grounded items cover lookup, comparison, and
calculation, using neighboring or semantically similar entries as
distractors. Calculations use boxed value--unit answers, whereas MCQs use
consistent labels and computational or conceptual tags. For numerical
items, $m$ records the target value, canonical unit, permitted conversions,
and comparison tolerance.

For the fault axis, we construct diagnostic cases by pairing
drive test measurements with engineering parameter records
from real network scenarios. Each question asks the model
to identify the cause of downlink-throughput degradation among eight fault
classes: excessive downtilt, overshooting, neighboring-cell effects,
overlapping coverage, frequent handovers, physical cell identity (PCI) interference, excessive
vehicle speed, and insufficient scheduled resources.

Only questions passing answer-validity, leakage, completeness, format, and
consistency checks are retained for subsequent CoT construction.

\subsubsection{Reasoning-Trace Generation and Verification}
We use multiple teacher models to generate reasoning traces under source-matched formats and diverse reasoning depths \cite{deepseekr1,guha2025openthoughts}. Following Eq.~\eqref{eq:acceptance_criterion}, only traces passing both
$C^{(a)}$ and $V^{(a)}$ are retained.

For the protocol axis, traces ground the answer in standards evidence,
compare candidate WGs, reject nearby alternatives, and return a
verifier-readable label~\cite{nikbakht2024tspecllm}. Verification checks the
WG label, candidate mapping, evidence support, and alternative rejection.

For the knowledge axis, traces justify the correct option and refute each
distractor~\cite{maatouk2025teleqna}. We retain concise and detailed traces
only when their reasoning is complementary rather than paraphrased.
Verification checks the selected option and the consistency of the
option-level analysis.

For the modeling axis, traces expose equations, substitutions, unit
conversions, code semantics, table evidence, and calculations supporting the
answer~\cite{wirelessmathlm2025,ezzakri2025teletables}.
For executable mathematical and code-grounded items, we translate the
calculations into python and rerun them. $C^{(a)}$ checks intermediate
values, units, and derivation consistency against the gold solution or
source evidence, while $V^{(a)}$ extracts the final answer and applies
symbolic equivalence, unit-aware numerical comparison, or option equality
according to the item type.

For the fault axis, we define a deterministic diagnostic rule set and execute it through five stages: feature extraction, metric calculation, threshold checking, rule selection, and fault class assignment. We then use this procedure to generate traces in a structured \textit{[Calculation]/[Diagnosis]/[Answer]} format. The calculation section presents the extracted features, computed metrics, and threshold comparisons. The diagnosis section states the fault cause determined by the selected rule, and the answer section maps that cause to the final option. During verification, we rerun the procedure. A trace is retained only when its calculations are correct and the replay matches both the expected rule identifier and fault class.

Finally, building on answer-verified reasoning-trace filtering
~\cite{deepseekr1}, we further apply prefix-continuation self-validation.
We truncate each trace before its conclusion and ask an independent model to
continue from the remaining prefix. The trace is retained only if the
continuation recovers the gold answer, confirming that the preceding
reasoning supports the conclusion.

\subsubsection{Failure-Driven Refinement and Expansion}
We use failed candidate responses in two complementary ways: to repair individual generation failures and to expand training data's coverage of recurring failure patterns. For the protocol, knowledge, and modeling axes, failure-mined knowledge injection
identifies the missing or misapplied fact, converts it into a compact
context entry, and conditions a new generation attempt on that entry.
The regenerated response is then evaluated by the same acceptance
pipeline defined in Eq.~\eqref{eq:acceptance_criterion}.

For the protocol axis, we categorize failures by standards family, WG responsibility boundary, interface ownership, specification type, and document relationship. The recovered standards knowledge is reintroduced during regeneration to help distinguish confusable WGs.

For the knowledge axis, we identify missing facts, long-tail concepts, and frequently confused distinctions. These facts are used both to regenerate failed responses and to guide the collection or construction of additional questions targeting recurring knowledge gaps.

For the modeling axis, we categorize failures by formula, telecom
subdomain, problem type, and failed reasoning step. Missing formulas,
assumptions, or calculation rules are supplied during regeneration,
while recurring failure categories guide further data construction.
We additionally instantiate new problems by varying the parameters of
verified calculation templates and recomputing all intermediate values 
and gold answers.

\subsubsection{Final Corpus}

After axis-specific verification, all examples undergo shared structural checks, duplicate removal, leakage filtering, difficulty grouping, and response-style mixing. To reduce exact-match benchmark leakage, we also remove any training example whose normalized question--answer pair exactly matches an item from our test set, the
GSMA Open Telco Leaderboard benchmark. The remaining examples are normalized into a common \texttt{\{system, user, assistant\}} format and stored with their axis and source tags.

\begin{figure}[t]
    \centering
    \includegraphics[width=0.9\columnwidth]
    {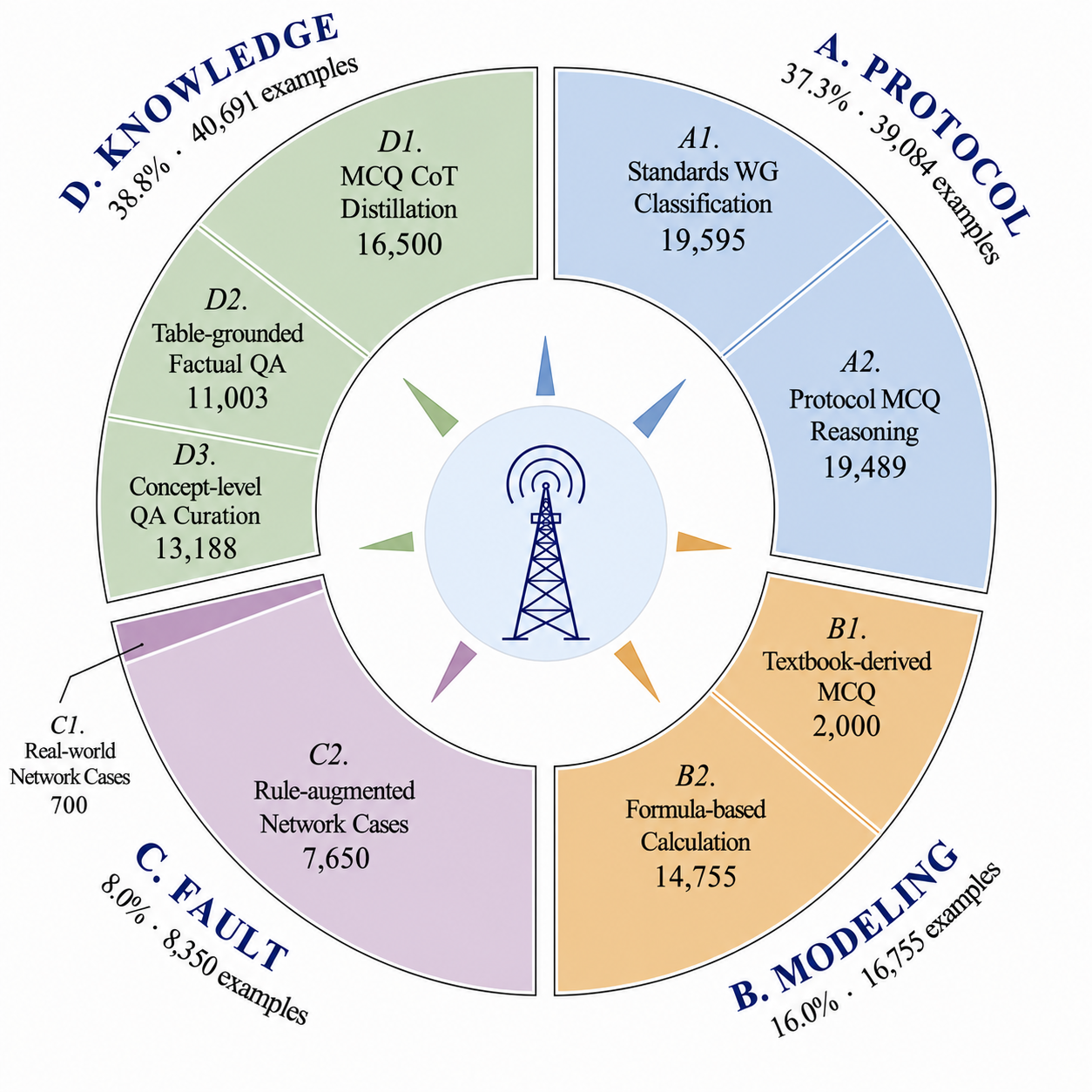}
    \caption{Composition of the $104{,}880$ examples of \textit{TelecomGPT-R1} corpus
    across four reasoning axes and source types.}
    \label{fig:data_composition}
\end{figure}

The final \textit{TelecomGPT-R1} corpus contains $104{,}880$ examples, comprising $92{,}380$ verified reasoning traces for SFT and $12{,}500$ prompts for RL. Fig.~\ref{fig:data_composition} shows their distribution across reasoning axes and source types. During RL, knowledge and modeling reuse their final-answer verifiers, while protocol and fault extend their verification procedures through the reward rubrics introduced in Sec.~\ref{sec:posttrain:reward}.

\subsection{Two-Stage Post-Training}
\label{sec:posttrain}

\begin{figure*}[!t]
    \centering
    \includegraphics[width=\textwidth]{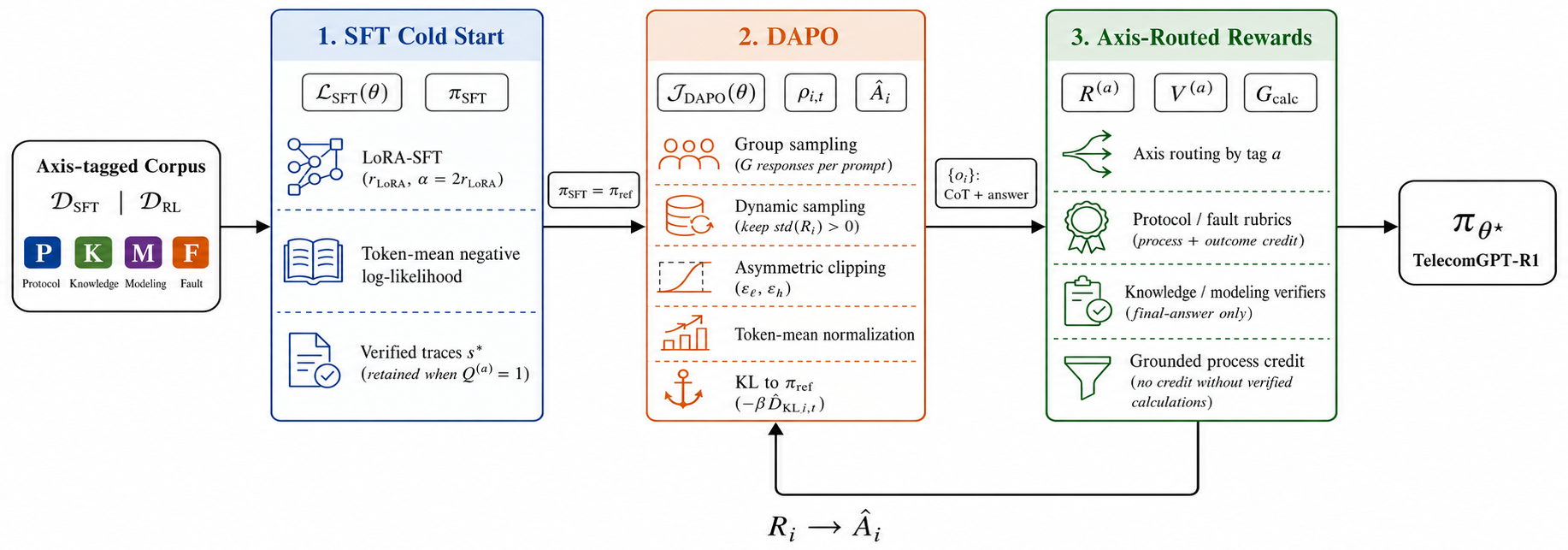}
    \caption{Two-stage post-training with axis-routed rewards.}
    \label{fig:recipe_reward}
\end{figure*}

Post-training consists of SFT followed by RL. We first perform LoRA-SFT on the verified multi-teacher reasoning
corpus. We then optimize the SFT policy using DAPO, a variant of
GRPO~\cite{deepseekr1,dapo2025}, with the axis-routed rewards defined in
Sec.~\ref{sec:posttrain:reward}. Fig.~\ref{fig:recipe_reward} summarizes the
complete procedure.

\subsubsection{SFT}
\label{sec:posttrain:sft}

We adapt the Qwen3.5 base model using LoRA adapters with rank
$r_{\mathrm{LoRA}}$ and scaling parameter
$\alpha=2r_{\mathrm{LoRA}}$. Given an SFT example
$(x,s^\star,y^\star,a,m)\sim\mathcal{D}_{\mathrm{SFT}}$, we minimize the
token-mean negative log-likelihood of the verified reasoning trace:
\begin{equation}
\label{eq:sft}
\mathcal{L}_{\mathrm{SFT}}(\theta)
=
-\mathbb{E}_{(x,s^\star)\sim\mathcal{D}_{\mathrm{SFT}}}\!
\left[
\frac{1}{|s^\star|}\!
\sum_{t=1}^{|s^\star|}\!
\log\pi_\theta\!
\left(
s_t^\star
\!\mid\!
x,s_{<t}^\star
\right)
\!\right]\!.
\end{equation}
Here, $s_t^\star$ is the $t$-th target token and $|s^\star|$ is the target
length. The resulting policy is denoted by $\pi_{\theta_{\mathrm{SFT}}}$ and is used
as both the RL initialization and the fixed reference policy.

\subsubsection{Axis-Routed Rewards}
\label{sec:posttrain:reward}

For each generated response $o$, the axis tag $a$ selects the corresponding
reward $R^{(a)}(o;y^\star,m)$. Protocol and fault use reward rubrics that
assign both final-answer and process credit. Knowledge and modeling use
direct final-answer verification. 
\paragraph{Protocol rubric}

Let $\hat{y}$ be the parsed WG prediction and $y^\star$ the gold
WG label. The protocol rubric asks four questions:

\begin{enumerate}
    \item Is the final WG prediction correct?
    \item Is the gold WG included among the candidates considered?
    \item Does the reasoning process identify the gold WG as a
    supported candidate but later reject it in the final comparison?
    \item Does the predicted group belong to the same family as the gold group?
\end{enumerate}

These questions produce the following reward:
\begin{equation}
\label{eq:reward_protocol}
\begin{aligned}
&R^{(\mathrm{protocol})}(o;y^\star,m)
=
\mathrm{clip}\Big(
V^{(\mathrm{protocol})}(o;y^\star,m) \\
&\quad+0.10\,\mathbbm{1}_{\mathrm{cand}}(o;y^\star) \\
&\quad-0.30\,\mathbbm{1}_{\mathrm{rej}}(o;y^\star) \\
&\quad+0.15\,\mathbbm{1}_{\mathrm{fam}}(\hat{y},y^\star),
\;-0.30,\;1.00
\Big).
\end{aligned}
\end{equation}

Here, $V^{(\mathrm{protocol})}=1$ when the final prediction is correct.
The indicator $\mathbbm{1}_{\mathrm{cand}}$ is one when the gold working
group is included among the candidates considered by the response.
The indicator $\mathbbm{1}_{\mathrm{rej}}$ is one when the reasoning process
identifies the gold group as a plausible or supported candidate but later
rejects it in the final comparison. This term penalizes overthinking that
reverses an otherwise correct intermediate decision.
The indicator $\mathbbm{1}_{\mathrm{fam}}$ is one when the predicted group
belongs to the same WG family as the gold label.

The rubric therefore rewards useful candidate reasoning while penalizing
cases in which the model reaches the correct candidate but reasons itself
away from it.

\paragraph{Fault rubric}

The fault rubric uses the deterministic diagnostic procedure introduced
during data construction. Let $\hat{y}$ and $y^\star$ be the predicted and
gold answer options, and let $\hat{d}$ and $d^\star$ be the predicted and
reference diagnoses.

The rubric first asks three questions about the supporting calculations:

\begin{enumerate}
    \item Are all metrics required by the diagnosis reported?
    \item Do the reported values match the deterministic calculations?
    \item Are the required threshold comparisons applied correctly?
\end{enumerate}

These questions form the binary calculation check
$G_{\mathrm{calc}}(o;d^\star,m)$. Diagnosis-level process credit is available
only when all three answers are yes. The rubric then asks:

\begin{enumerate}
    \setcounter{enumi}{3}
    \item Does the supported diagnosis match the reference diagnosis?
    \item Does the final answer option match the gold option?
\end{enumerate}

The resulting reward is
\begin{equation}
\label{eq:reward_fault}
\begin{aligned}
&R^{(\mathrm{fault})}(o;y^\star,m)
=
\mathrm{clip}\Big(
\mathbbm{1}[\hat{y}=y^\star] \\
&\quad+\lambda_{\mathrm{proc}}\,
G_{\mathrm{calc}}(o;d^\star,m)\,
\mathbbm{1}[\hat{d}=d^\star],
\;0,\;1
\Big),
\end{aligned}
\end{equation}
where $\lambda_{\mathrm{proc}}=0.5$.

Let $\mathcal{J}(d^\star)$ be the set of metrics required by the reference
diagnosis. For each metric $j$, the parser extracts the reported value
$\hat{v}_j$ and the stated result $\hat{b}_j\in\{0,1\}$ of the corresponding
threshold comparison. The calculation check is
\begin{equation}
\label{eq:fault_calculation_check}
\begin{aligned}
G_{\mathrm{calc}}(o;d^\star,m)
=
\prod_{j\in\mathcal{J}(d^\star)}
\Big(
&\mathbbm{1}[\hat{v}_j\text{ is reported}]
\cdot \mathrm{close}(\hat{v}_j,v_j^\star) \\
&\cdot \mathbbm{1}[\hat{b}_j=b_j^\star]
\Big),
\end{aligned}
\end{equation}
where $v_j^\star$ is recomputed from the input metadata $m$, and
$b_j^\star$ is the correct result of applying the corresponding threshold
condition to $v_j^\star$.

Metric agreement is defined as
\begin{equation}
\label{eq:close}
\mathrm{close}(\hat{v},v^\star)
=
\begin{cases}
\mathbbm{1}
\left[
|\hat{v}|
\le
\tau_{\mathrm{abs}}
\right],
& v^\star=0,\\[3pt]
\mathbbm{1}
\left[
|\hat{v}-v^\star|
\le
\tau_{\mathrm{rel}}|v^\star|
\right],
& v^\star\neq0,
\end{cases}
\end{equation}
where $\tau_{\mathrm{rel}}=0.05$ and $\tau_{\mathrm{abs}}=0.5$.

The calculations do not receive reward independently. They determine only
whether the diagnosis is eligible for process credit. Mentioning the
reference diagnosis, listing many possible causes, or reporting unrelated
metrics therefore receives no process credit unless all required
calculations and threshold comparisons support that diagnosis.

\paragraph{Modeling reward}

The modeling reward uses the final symbolic--numeric verifier defined in Eq.~\eqref{eq:verifier}:
\begin{equation}
\label{eq:reward_modeling}
R^{(\mathrm{modeling})}(o;y^\star,m)
=
V^{(\mathrm{modeling})}(o;y^\star,m).
\end{equation}
The verifier extracts the boxed answer and checks symbolic equivalence or
numerical agreement under the corresponding tolerance.

\paragraph{Knowledge reward}

The knowledge reward uses the verifier defined in Eq.~\eqref{eq:verifier} with parsed answer equality:
\begin{equation}
\label{eq:reward_knowledge}
R^{(\mathrm{knowledge})}(o;y^\star,m)
=
V^{(\mathrm{knowledge})}(o;y^\star,m).
\end{equation}
Reasoning-trace quality for this axis is controlled during data construction,
while RL checks the final MC answer.

\subsubsection{DAPO Optimization}
\label{sec:posttrain:dapo}

We instantiate the RL objective in Eq.~\eqref{eq:posttraining_objective} using
DAPO, a variant of GRPO~\cite{deepseekr1,dapo2025}. 
At each RL iteration, $\pi_{\theta_{\mathrm{old}}}$ denotes the frozen
behavior policy used to generate the current rollout batch. The initial $\theta_{\mathrm{old}}$ is $\theta_{\mathrm{SFT}}$, and before each subsequent batch,
it is replaced by the latest policy parameters.
For each
$(x,y^\star,a,m)\in\mathcal{D}_{\mathrm{RL}}$, the behavior policy samples
$G$ responses and evaluates them with the axis-routed reward:
\begin{equation}
\label{eq:dapo_sampling}
o_i\sim\pi_{\theta_{\mathrm{old}}}(\cdot\!\mid\!x),
\;\;
R_i=R^{(a)}(o_i;y^\star\!,m),
\;\;
i=1,\ldots,G.
\end{equation}

DAPO retains only groups that provide a nonzero relative learning signal:
\begin{equation}
\label{eq:dynamic_buffer}
\mathcal{B}_{\mathrm{dyn}}
=
\left\{
\left(x,\{o_i,R_i\}_{i=1}^{G}\right):
\operatorname{std}_{i=1}^{G}(R_i)>0
\right\}.
\end{equation}
If all responses receive the same reward, their group-relative advantages
are zero and the group cannot contribute to the policy update. Dynamic
sampling filters such groups and continues sampling until the effective
training batch is filled.

For each retained group, the normalized advantage and token-level importance
ratio are
\begin{equation}
\label{eq:dapo_statistics}
\hat{A}_i
\!=\!
\frac{
R_i\!-\!\operatorname{mean}_{j=1}^{G}R_j
}{
\operatorname{std}_{j=1}^{G}R_j\!+\!\eta
},
\;
\rho_{i,t}(\theta)
\!=\!
\frac{
\pi_\theta(o_{i,t}\!\mid\!x,o_{i,<t})
}{
\pi_{\theta_{\mathrm{old}}}(o_{i,t}\!\mid\!x,o_{i,<t})
},
\end{equation}
where $\eta>0$ ensures numerical stability. A positive $\hat{A}_i$
increases the probability of the sampled response, whereas a negative value
suppresses it. The same response-level advantage is applied to all tokens in
$o_i$.

We optimize the following KL-regularized DAPO objective:
\begin{equation}
\label{eq:dapo_objective}
\begin{aligned}
\mathcal{J}_{\mathrm{DAPO}}(\theta)
={}&
\mathbb{E}_{\mathcal{B}_{\mathrm{dyn}}}
\Bigg[
\frac{1}{\sum_i |o_i|}
\sum_i\sum_t
\Bigg( \\
&\quad
\min\Big[
\rho_{i,t}\hat{A}_i,\;
\operatorname{clip}\!\big(
\rho_{i,t},
1{-}\varepsilon_\ell,
1{+}\varepsilon_h
\big)\hat{A}_i
\Big] \\
&\quad
-\beta\widehat{D}_{\mathrm{KL},i,t}
\Bigg)
\Bigg],
\end{aligned}
\end{equation}
where asymmetric clipping, with $\varepsilon_h>\varepsilon_\ell$, provides
additional room for increasing promising low-probability tokens while
constraining unstable policy changes. The token-level KL estimator relative
to $\pi_{\mathrm{ref}}=\pi_{\theta_{\mathrm{SFT}}}$ is
\begin{equation}
\label{eq:token_kl}
\widehat{D}_{\mathrm{KL},i,t}
=
\xi_{i,t}\!-\!\log\xi_{i,t}\!-\!1,
\;\;
\xi_{i,t}
=
\frac{
\pi_{\mathrm{ref}}(o_{i,t}\!\mid\!x,o_{i,<t})
}{
\pi_\theta(o_{i,t}\!\mid\!x,o_{i,<t})
}.
\end{equation}

The optimizer minimizes
$\mathcal{L}_{\mathrm{DAPO}}=-\mathcal{J}_{\mathrm{DAPO}}$.
Dynamic sampling preserves effective reward variation, asymmetric clipping
stabilizes probability updates, and normalization by the total number of
generated tokens prevents per-response averaging from diluting the learning
signal of long CoT responses.

\section{Main Results}\label{sec:main_results}

\begin{table*}[t]
    \renewcommand{\arraystretch}{1.15}
    \caption{Per-dataset and mean accuracy (\%) on the GSMA Open Telco Leaderboard.}
    \label{tab:main}
    \centering
    \footnotesize
    \begin{tabular}{l c c c c c c c|c}
    \hline\hline
    Model & 3GPP-TSG & ORANBench & srsRANBench & TeleLogs & TeleMath & TeleQnA & TeleTables\textsuperscript{$\dagger$} & Mean \\
    \hline
    \multicolumn{9}{l}{\emph{Closed source operator-internal}} \\
    AT\&T OTel-LLM-8.3B-QnA & 81.4 & \textbf{94.1} & 89.7 & 96.3 & \textbf{87.4} & \textbf{91.2} & 61.8 & 86.0 \\
    China Telecom TeleLLM & 78.6 & 77.4 & 82.0 & 85.2 & 71.2 & 89.4 & 46.6 & 75.8 \\
    SoftBank LTM & 68.4 & 82.0 & 83.1 & 73.4 & 81.5 & 81.9 & 44.7 & 73.6 \\
    \hline
    \multicolumn{9}{l}{\emph{Closed source general-purpose frontier}} \\
    Gemini-3.1-Pro-Preview~\cite{google2025gemini} & 70.0 & 86.0 & 84.7 & 82.0 & 73.0 & 85.2 & 48.0 & 75.6 \\
    Claude-Opus-4.6~\cite{anthropic2025claude45} & 66.0 & 90.0 & 84.7 & 70.0 & 75.0 & 84.4 & 43.0 & 73.3 \\
    GPT-5~\cite{openai2025gpt5} & 58.0 & 86.0 & 81.3 & 78.0 & 79.0 & 83.8 & 37.0 & 71.9 \\
    Kimi-K2.5 & 57.0 & 82.7 & 84.7 & 60.0 & 76.0 & 83.6 & 42.0 & 69.4 \\
    o3 & 59.0 & 84.7 & 78.7 & 72.0 & 74.0 & 83.4 & 34.0 & 69.4 \\
    Grok-4-fast~\cite{xai2025grok4} & 52.0 & 83.3 & 84.0 & 75.0 & 66.0 & 83.6 & 35.0 & 68.4 \\
    \hline
    \multicolumn{9}{l}{\emph{Open source general-purpose}} \\
    DeepSeek-V3 & 49.0 & 79.3 & 80.0 & 40.0 & 57.0 & 81.6 & 28.0 & 59.3 \\
    GPT-OSS-120B & 32.0 & 80.0 & 84.0 & 45.0 & 57.0 & 79.9 & 30.0 & 58.3 \\
    LLaMA-3.3-70B-Instruct & 52.0 & 76.0 & 85.3 & 18.0 & 45.0 & 77.3 & 29.0 & 54.7 \\
    Qwen2.5-72B-Instruct & 45.1 & 74.7 & 79.5 & 27.3 & 45.5 & 76.5 & 29.2 & 54.0 \\
    Qwen3.5-4B~\cite{qwen35} & 26.5 & 67.1 & 75.7 & 35.7 & 11.4 & 76.3 & 77.7 & 52.9 \\
    Qwen3.5-9B~\cite{qwen35} & 32.4 & 68.4 & 77.0 & 37.9 & 15.4 & 77.5 & 67.3 & 53.7 \\
    Qwen3.5-27B~\cite{qwen35} & 45.3 & 76.4 & 79.2 & 52.4 & 29.8 & 81.3 & 84.9 & 64.2 \\
    Gemma3-27B & 39.6 & 71.6 & 80.6 & 16.0 & 40.7 & 71.3 & 33.3 & 50.4 \\
    \hline
    \multicolumn{9}{l}{\emph{Open source telecom (ours, SFT-only)}} \\
    TelecomGPT-R1-4B  & 71.2 & 78.1 & 81.9 & 49.1 & 61.4 & 80.8 & 79.8 & 71.8 \\
    TelecomGPT-R1-9B  & 74.4 & 84.1 & 81.9 & 55.2 & 72.0 & 84.9 & 87.5 & 77.1 \\
    TelecomGPT-R1-27B & 72.0 & 89.3 & 84.7 & 64.8 & 77.4 & 85.4 & 90.0 & 80.5 \\
    \hline
    \multicolumn{9}{l}{\emph{Open source telecom (ours, SFT+DAPO)}} \\
    TelecomGPT-R1-4B  & 75.4 & 81.6 & 87.0 & 85.5 & 66.8 & 82.5 & 79.5 & 79.8 \\
    TelecomGPT-R1-9B  & 77.9 & 85.8 & 85.0 & 87.5 & 75.2 & 85.7 & 86.9 & 83.4 \\
    TelecomGPT-R1-27B & \textbf{84.2} & 91.4 & \textbf{91.3} & \textbf{99.0} & 82.4 & \textbf{91.2} & \textbf{88.0} & \textbf{89.6} \\
    \hline
    \end{tabular}
    \vspace{2pt}
    \begin{minipage}{0.98\textwidth}
    \raggedright
    \scriptsize
    \textsuperscript{$\dagger$}The \textit{TelecomGPT-R1} results on TeleTables follow the TeleTables-Easy evaluation setting in~\cite{ezzakri2025teletables}.
    \end{minipage}
\end{table*}

We report the performance of \textit{TelecomGPT-R1} on the seven-benchmark GSMA Open Telco Leaderboard, together with the evaluation and implementation settings.

\subsection{Evaluation Setup}
\label{sec:exp:protocol}

We evaluate all public test examples without subsampling~\cite{telecomgpt2024}. 3GPP-TSG is the protocol-axis benchmark whose 16-way task classifies stems into 3GPP WGs \cite{gsma2026open_telco_evals}. ORANBench sits on the knowledge axis and probes O-RAN interface concepts through MCQ \cite{gajjar2025oran}. srsRANBench is the modeling-axis benchmark that scores code understanding via srsRAN source-code MC \cite{gajjar2025oransight}. TeleLogs is the fault-axis benchmark that asks for fault-class inference over engineering-log traces \cite{sana2025reasoning}. TeleMath is the modeling-axis benchmark that scores numerical and symbolic problem solving on wireless-engineering derivations \cite{colle2026telemath}. TeleQnA sits on the knowledge axis and evaluates 5-option knowledge MC \cite{maatouk2025teleqna}. TeleTables is the knowledge-axis benchmark that requires structured-table lookup MC over standards tables \cite{ezzakri2025teletables}. Inference uses greedy decoding under a 4{,}096-token prompt window and a per-completion budget of up to 8{,}192 response tokens for derivation tasks. Verifiers are task-matched: unit-tolerant numerical equality for TeleMath, letter-choice match for the four MC tasks, and stage-by-stage rubric scoring for TeleLogs and 3GPP-TSG.

\subsection{Implementation and Hyperparameters}
\label{sec:exp:impl}

We use Qwen3.5-4B/9B/27B as the base models. SFT uses 92{,}380 multi-teacher examples with LoRA ($r=32$, $\alpha=64$, dropout $0.05$), AdamW, a learning rate of $2\times10^{-5}$, a global batch size of 256, and two epochs. DAPO uses dynamic sampling, token-level loss aggregation, asymmetric clipping $(\varepsilon_\ell,\varepsilon_h)=(0.20,0.28)$, and a fixed SFT reference with $\beta=0.001$. Responses are capped at $L_{\max}=8{,}192$ tokens, with overflow rollouts removed by the reward manager. Protocol and fault use verifier-grounded rubric rewards based on decision consistency and deterministic replay, respectively. Training and evaluation run on $8\times$ NVIDIA H200 GPUs with vLLM. Reproduction scripts are released, and Sec.~\ref{sec:posttrain} provides the formal objectives.

\subsection{Comparison to Baselines}\label{sec:main_results:comparison}

Table~\ref{tab:main} and Fig.~\ref{fig:radar} compare five model groups: operator-internal telecom models, closed-source and open-source general-purpose models, SFT-only variants, and the full \textit{TelecomGPT-R1} family. Public models use identical greedy decoding, while operator-internal scores follow the leaderboard snapshot.
For TeleTables, our \textit{TelecomGPT-R1} evaluations follow the TeleTables-Easy setting in~\cite{ezzakri2025teletables}.

\textit{TelecomGPT-R1-27B} achieves the best mean of 89.6\%, outperforming the three representative closed source operator-internal baselines: AT\&T OTel-LLM-8.3B-QnA at 86.0\%, China Telecom TeleLLM at 75.8\%, and SoftBank LTM at 73.6\%. Against the strongest, AT\&T OTel-LLM-8.3B-QnA, R1-27B leads by +3.6 percentage points (pp): it improves on 3GPP-TSG, srsRANBench, TeleLogs, and TeleTables, ties on TeleQnA, and trails on ORANBench and TeleMath. The largest margin, +26.2 pp on TeleTables, reflects the gain from structured-evidence reasoning.

Against the strongest closed source general-purpose frontier models, Gemini-3.1-Pro-Preview~\cite{google2025gemini}, Claude-Opus-4.6~\cite{anthropic2025claude45}, and GPT-5~\cite{openai2025gpt5}, R1-27B leads by +14.0, +16.3, and +17.7 pp, respectively. Among open generalists, Qwen3.5-27B~\cite{qwen35} leads at 64.2\%, the base backbone of \textit{TelecomGPT-R1}, ahead of DeepSeek-V3-685B at 59.3\% and GPT-OSS-120B at 58.3\%. R1-27B gains +30.3 pp over DeepSeek-V3-685B while using $\sim$1/25 the parameter count, indicating that telecom-specific reasoning matters more than scale.

The SFT-only comparison quantifies the effect of the full post-training pipeline: mean scores rise from 71.8\%, 77.1\%, and 80.5\% to 79.8\%, 83.4\%, and 89.6\% for the 4B, 9B, and 27B models. These consistent gains confirm the value of axis-aware data construction and rubric-guided post-training.

\section{Analysis}\label{sec:analysis}

We now analyze what is required to train a broadly capable telecom reasoner. Rather than presenting isolated ablations, we organize the analysis around four central design questions: how to cold start the model before RL, how to train one unified policy across heterogeneous telecom sources, how to construct informative yet robust rewards, and how the recipe scales with model and data size.
\subsection{On Cold Start of Telecom Reasoners}
\label{sec:analysis:coldstart}

\subsubsection{SFT is necessary before RL}

\begin{figure}[t]
\centering
\includegraphics[width=\columnwidth]{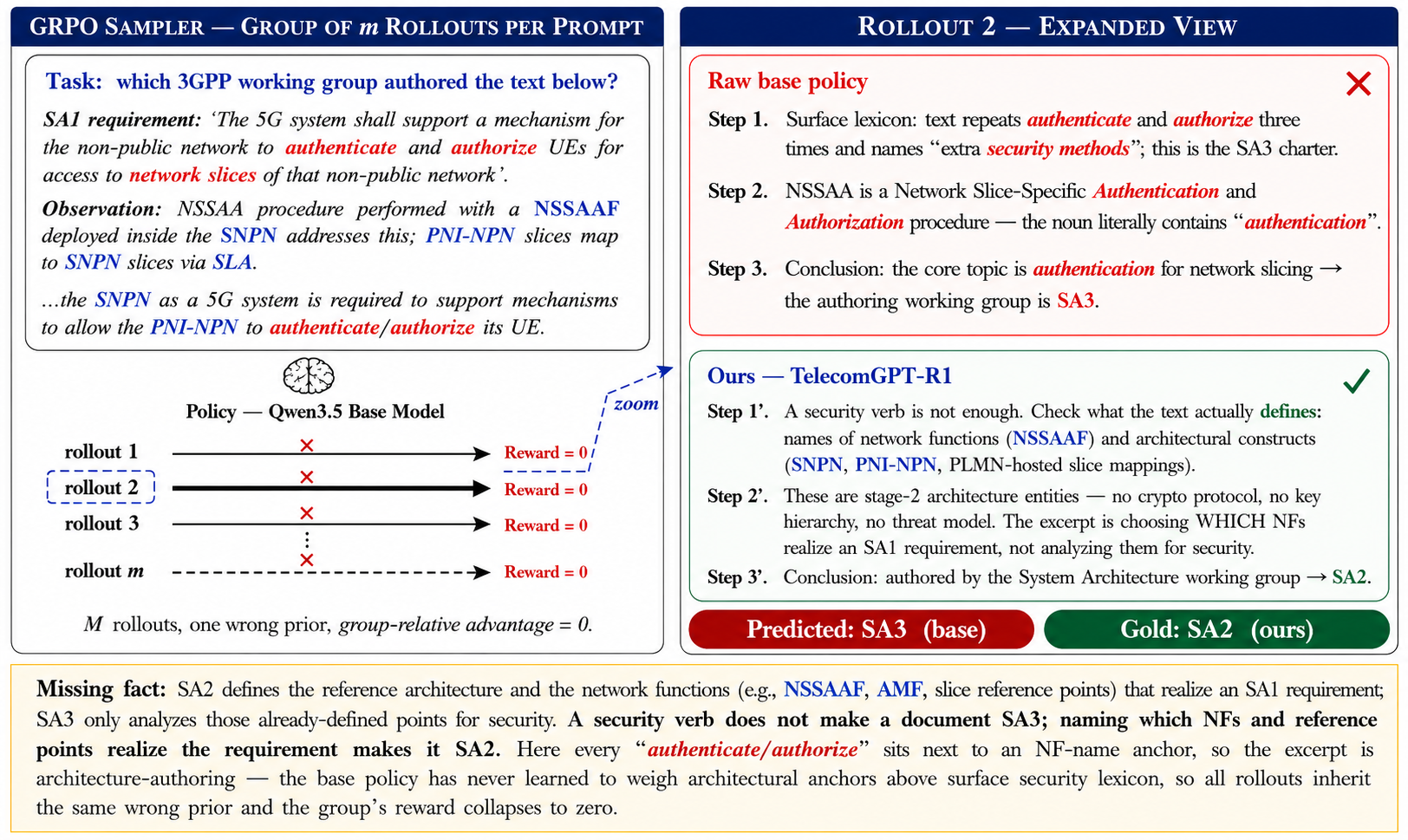}
\caption{Cold start failure on a protocol example.}
\label{fig:coldstart_wrong_fact}
\end{figure}

RL provides little useful signal when the initial policy fails to generate any correct, telecom-grounded response within a rollout group. An important cold start barrier is that the initial policy may generate fluent responses while relying on incorrect telecom knowledge. Fig.~\ref{fig:coldstart_wrong_fact} shows a representative protocol axis failure. During RL training, the base Qwen3.5 policy associates the terms ``authenticate'' and ``authorize'' with System Aspects (SA3) and repeatedly predicts SA3 across all rollouts, while missing that the NSSAAF, SNPN, and PNI-NPN context identifies SA2 as the responsible WG. The sampled responses differ in wording but consistently rely on the same incorrect SA2--SA3 responsibility boundary.

This failure especially limits group-relative RL. If the policy cannot generate any factually valid trajectory for a prompt $x$, all sampled rollouts receive the same zero reward. The group reward variance is then zero, leaving no positive response to reinforce through relative advantages. Increasing the number of rollouts may diversify their surface forms, but in this example the responses continue to reproduce the same incorrect factual premise. SFT is therefore needed to make correct, telecom-grounded responses more likely before RL can reliably reinforce them.

Table~\ref{tab:main} quantifies this cold start effect. On the 27B backbone, performance increases from $64.2\%$ for the base model to $80.50\%$ after SFT and $89.64\%$ after subsequent RL. SFT contributes a $16.3$ pp improvement, while RL contributes a further $9.1$ pp. Moreover, Fig.~\ref{fig:model_size_scaling} shows that RL-only training improves the base model but consistently underperforms SFT-only and SFT+RL across model sizes. Thus, SFT provides the factual support and policy initialization required for effective RL.

\subsubsection{Even strong teachers fail to provide valid CoT coverage for long-tail telecom questions}

We next investigate how to construct accurate and useful CoT traces for telecom SFT, focusing on two questions: how to correct missing telecom knowledge in teacher-generated traces and how to match each trace to the reasoning structure of its source. 

Although general domain post-training commonly distills reasoning traces directly from strong teachers~\cite{guha2025openthoughts} such as GPT family models, we find that direct distillation remains insufficient for telecom CoT data generation.  
Taking the protocol axes as an example, even when the teacher receives detailed instructions and the gold answer, direct distillation of teachers such as GPT-5.4 and DeepSeek-V4-Pro produces at least one verifier-valid trace for only approximately 50\% of the evaluated questions, as shown in Table~\ref{tab:teacher_injection}. Here, coverage denotes the percentage of questions for which at least one valid trace is obtained. The uncovered questions typically reflect factual rather than stylistic failures. These failures involve missing or confused standards facts, including WG responsibility boundaries, defining versus referenced specifications, and document ownership. Consequently, direct teacher
generation leaves many questions without any verifier-valid reasoning
trace. Simply filtering invalid generations would therefore discard a
substantial fraction of the intended training questions, particularly
those involving long-tail or easily confused telecom knowledge.

This motivates our failure-mined knowledge injection pipeline, which identifies the missing or misapplied fact behind each failed trace, converts it into a compact knowledge entry, and provides it as additional context for a new generation attempt. This targeted evidence helps the teacher produce a trace that reaches the gold answer through verifiable telecom reasoning. 
We retain the regenerated trace only if it passes three checks. First, gold-answer verification confirms that the trace reaches the correct conclusion. Second, leakage filtering removes traces that reveal or assume the answer before presenting the supporting evidence. Third, prefix-continuation self-validation tests whether an independent model inference run can produce the correct answer from a truncated reasoning prefix, ensuring that the trace contains sufficient evidence rather than merely appending the gold answer at the end.

Failure-mined knowledge injection increases coverage on the same 3GPP question set from approximately $50\%$ to $100\%$. That is, every question yields at least one verifier-valid regenerated trace; the result does not imply that every generation attempt succeeds. After leakage filtering and prefix-continuation self-validation, $99.2\%$ of the questions retain at least one trace in the final SFT corpus. The pipeline therefore converts most initially uncovered questions into verified supervision rather than simply discarding them.

\begin{table}[t]
\centering
\caption{Effect of failure mined knowledge injection on Protocol CoT trace curation.}
\label{tab:teacher_injection}
\scriptsize
\setlength{\tabcolsep}{4pt}
\begin{tabular}{@{}>{\raggedright\arraybackslash}p{0.31\columnwidth}
                >{\centering\arraybackslash}p{0.18\columnwidth}
                >{\raggedright\arraybackslash}p{0.43\columnwidth}@{}}
\hline\hline
Curation stage & Coverage & Interpretation \\
\hline
Direct teacher generation
& $\sim$50\%
& \% of questions with at least one verifier valid directly generated trace  \\

After knowledge injection
& 100\%
& \% of questions with at least one verifier valid regenerated trace \\

Final SFT corpus
& 99.2\%
& \% of data passing both leakage and prefix continuation checks \\
\hline
\end{tabular}
\end{table}

\subsubsection{Source-matched CoT and teacher diversity matter}
\begin{table*}[t]
\centering
\caption{Source-matched CoT and teacher diversity ablation on TeleMath and TeleLogs with 9B backbone.}
\label{tab:source_cot_teacher_diversity}
\footnotesize
\setlength{\tabcolsep}{6pt}
\renewcommand{\arraystretch}{1.15}
\begin{tabular}{l|l|l|c c}
\hline\hline
Stage & SFT corpus & Design keywords & TeleMath & TeleLogs \\
\hline
\multirow{3}{*}{\emph{SFT only}}
 & No CoT                            & Answer-only target, no reasoning trace                    & 44.0                      & 53.0 \\
 & Generic CoT                           & Single generic teacher, source-blind CoT                  & 63.8                         & 59.0   \\
 &  Ours               & Axis-matched 4-quadrant generators, multi-teacher mixture & 72.0                       & 55.2 \\
\hline
\multirow{2}{*}{\emph{DAPO}}
 & Generic CoT SFT model as init                 & Single generic teacher, source-blind CoT                  & 72.0                         & 72.9   \\
 & Our SFT model as init & Axis-matched 4-quadrant generators, multi-teacher mixture & \textbf{75.2}                       & \textbf{87.5} \\
\hline
\end{tabular}
\end{table*}

Beyond injecting missing telecom facts, SFT traces must also match the evidence structure of each problem type. Telecom tasks differ not only in topic but also in the form of reasoning needed to justify an answer. 
Thus, a source-blind CoT generator can impose one generic explanation style across these fundamentally different verifier structures, weakening the correspondence between the generated reasoning and the evidence.

We therefore use source-matched CoT generators. Protocol traces emphasize WG ownership, specification families, and standards evidence. Modeling traces expose equation derivations, code semantics, or table-cell grounding. Fault traces follow the structured \textit{[Calculation]/[Rules]/[Answer]} format in diagnostic procedure. Knowledge traces remain shorter and emphasize precise concept disambiguation. This design provides supervision for both final-answer correctness and the intermediate evidence structures expected for each task.
We also found that teacher diversity further improves SFT quality by reducing systematic teacher-specific errors. 

Table~\ref{tab:source_cot_teacher_diversity} confirms the benefit of this design. On TeleMath, source-matched multi-teacher CoT raises SFT accuracy to $72.0\%$, compared with $63.8\%$ for generic CoT and $44.0\%$ for answer-only supervision. On TeleLogs, SFT+RL reaches $87.5\%$ from the source-matched initialization, compared with $72.9\%$ from generic CoT. Thus, combining accurate domain knowledge with source-appropriate, teacher-diverse traces strengthens SFT and substantially improves downstream RL trainability.

\subsection{On Multi-Source Training of Telecom Reasoners}
\label{sec:analysis:multisource}

A useful telecom reasoner must handle diverse tasks that arise in real telecom engineering workflows, such as standards interpretation, knowledge lookup, code and configuration understanding, mathematical modeling, table reasoning, and fault diagnosis. A deployed model may encounter these task and evidence types across different requests and should retain broad competence without requiring a separate specialist for each source. We therefore study whether competence learned on one telecom source transfers to others, and whether one unified policy can retain strong performance across sources.

\subsubsection{Single-source training leads to narrow specialization}
\begin{figure}[t]
    \centering
    \begin{subfigure}{0.98\columnwidth}
        \centering
        \includegraphics[width=\linewidth]{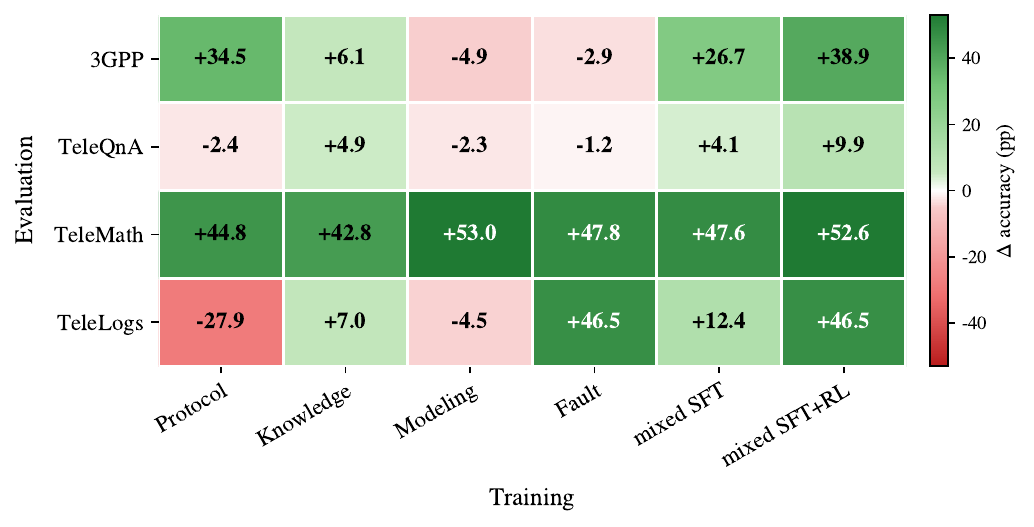}
        \caption{$\Delta$ accuracy for \textit{TelecomGPT-R1-27B} .}
        \label{fig:cross_source_27b}
    \end{subfigure}\\[4pt]
    \begin{subfigure}{0.98\columnwidth}
        \centering
        \includegraphics[width=\linewidth]{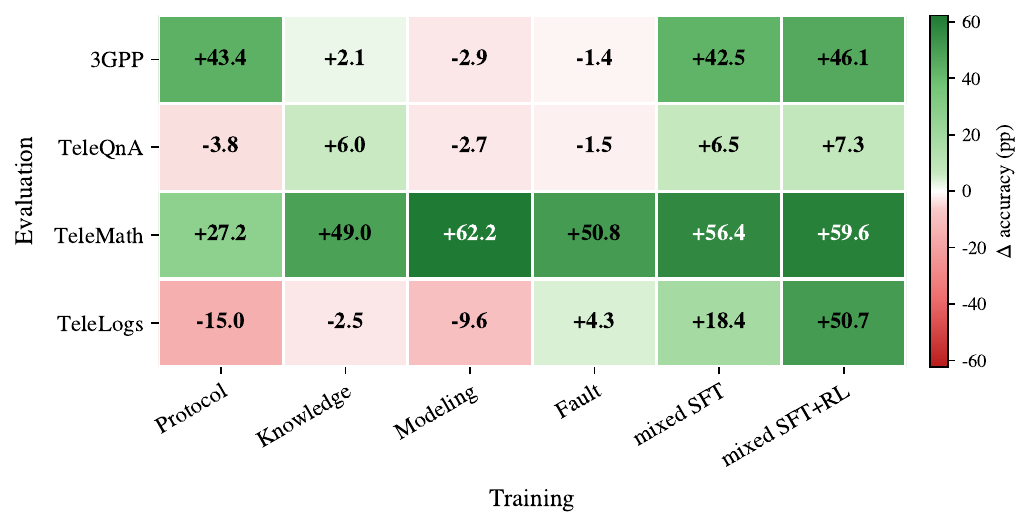}
        \caption{$\Delta$ accuracy for \textit{TelecomGPT-R1-9B}.}
        \label{fig:cross_source_9b}
    \end{subfigure}
    \caption{Cross-source transfer heatmap for both R1 variants.}
    \label{fig:cross_source_transfer}
\end{figure}

We perform a controlled transfer experiment, training models on each individual telecom axis and evaluating them on one representative benchmark from each axis. As shown by the transfer matrices for the 9B and 27B backbones in Fig.~\ref{fig:cross_source_transfer}, gains are highly localized in the source axis used for training. For example, protocol-only training improves standards-related tasks but provides little benefit to modeling or fault diagnosis, while some off-source capabilities remain unchanged or degrade.

The localized gains show that exposure to one telecom source does not by itself produce broad cross-source competence. Single-source training adapts the model to the response format, evidence-selection pattern, and reasoning procedure of that source, with limited transfer to others. Because reasoning patterns across axes differ substantially, specialization on one source transfers weakly and may interfere with others.

\subsubsection{Telecom sources induce different reasoning lengths and behaviors}
\begin{figure}[t]
\centering
\includegraphics[width=0.95\columnwidth]{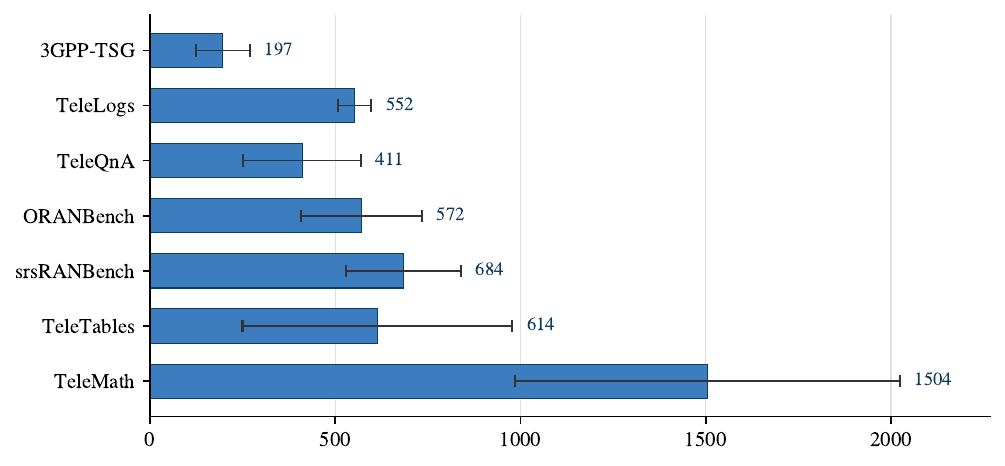}
\caption{Mean rollout response length per benchmark.}
\label{fig:reasoning_length_by_axis}
\end{figure}

As shown in Fig.~\ref{fig:reasoning_length_by_axis}, telecom sources induce distinct response patterns. Mathematical and fault tasks produce longer traces containing derivations, rule activations, and metric comparisons. Protocol tasks use medium-length traces to compare WGs and specification families, while knowledge, O-RAN, and table tasks favor shorter evidence extraction. The mean response length differs by up to $7.63\times$ between mathematical and protocol reasoning. Although length alone does not determine cross-source transfer, this variation shows that a unified policy must adapt its reasoning procedure to each source rather than impose one response pattern across tasks.


\subsubsection{Diverse task mixing improves performance across sources}

Given the limited transfer from single-source training, we examine whether mixed-source training can combine task-specific gains in one policy. As shown by the \textit{Merged SFT} results in Fig.~\ref{fig:cross_source_transfer}, diverse SFT largely removes negative off-source transfer and improves all four axes on both backbones. Across 3GPP, TeleQnA, TeleMath, and TeleLogs, the 27B model gains $(26.7, 4.1, 47.6, 12.4)$ pp after SFT, while the 9B model gains $(42.5, 6.5, 56.4, 18.4)$ pp. RL further increases these gains to $(38.9, 9.9, 52.6, 46.5)$ pp and $(46.1, 7.3, 59.6, 50.7)$ pp, respectively.
These results show that source-diverse training can consolidate largely distinct task-specific capabilities within one policy without requiring a separate specialist for each source.

\subsubsection{Multi-source RL benefits from non-uniform sampling}

\begin{figure}[t]
    \centering
    \includegraphics[width=\columnwidth]{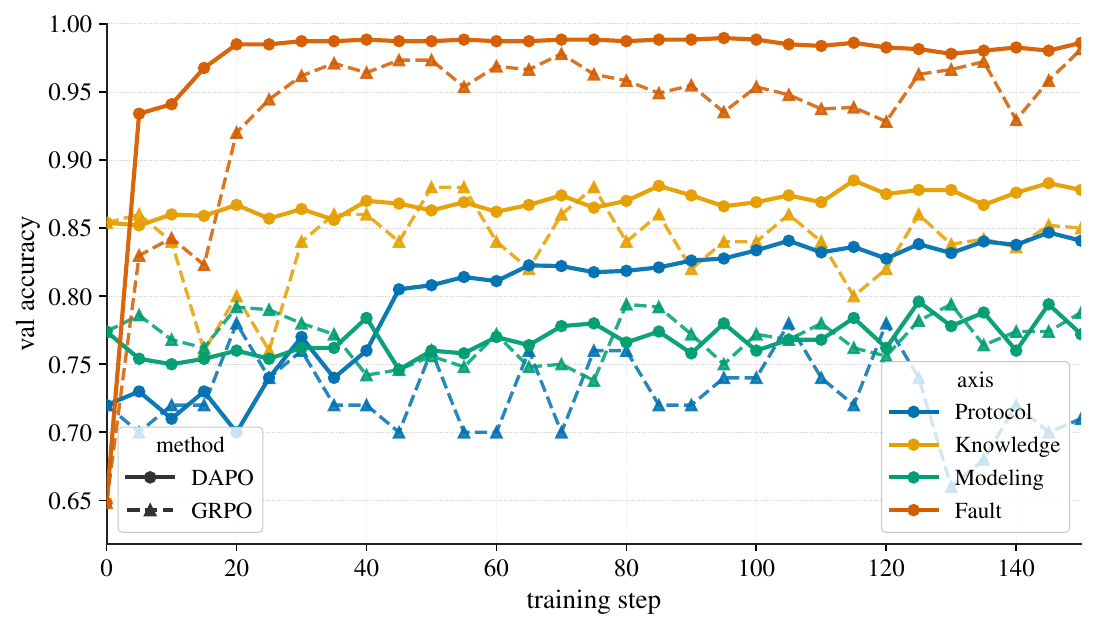}
    \caption{DAPO vs vanilla GRPO training dynamics for \textit{TelecomGPT-R1-27B} .}
    \label{fig:dapo_vs_grpo_multisource}
\end{figure}

During multi-source RL, the four telecom axes improve at different rates. As shown in Fig.~\ref{fig:dapo_vs_grpo_multisource}, fault reasoning approaches saturation early, whereas protocol reasoning improves over a longer period. Uniform sampling may continue allocating updates to prompts whose rollout groups no longer contain reward variation, while other prompts still provide informative comparisons.

DAPO addresses this imbalance by filtering rollout groups without reward variation and retaining prompts that still provide learning signals. Fig.~\ref{fig:dapo_vs_grpo_multisource} compares DAPO with vanilla GRPO on \textit{TelecomGPT-R1-27B}. Both methods start from the same SFT checkpoint and use the same source-specific rewards. The clearest difference appears on the protocol axis. DAPO produces steadier and higher protocol accuracy than GRPO, while allowing the fault axis to reach high accuracy within the first few training steps and maintain stable performance afterward. 

These results show that multi-source RL must account for differences in task difficulty and convergence speed. By focusing updates on informative rollout groups, DAPO improves training efficiency and stability across heterogeneous axes, especially the harder axes. 

\subsection{On Reward Design of Telecom Reasoners}
\label{sec:analysis:reward}
\begin{figure}
    \centering
    \includegraphics[width=1\linewidth]{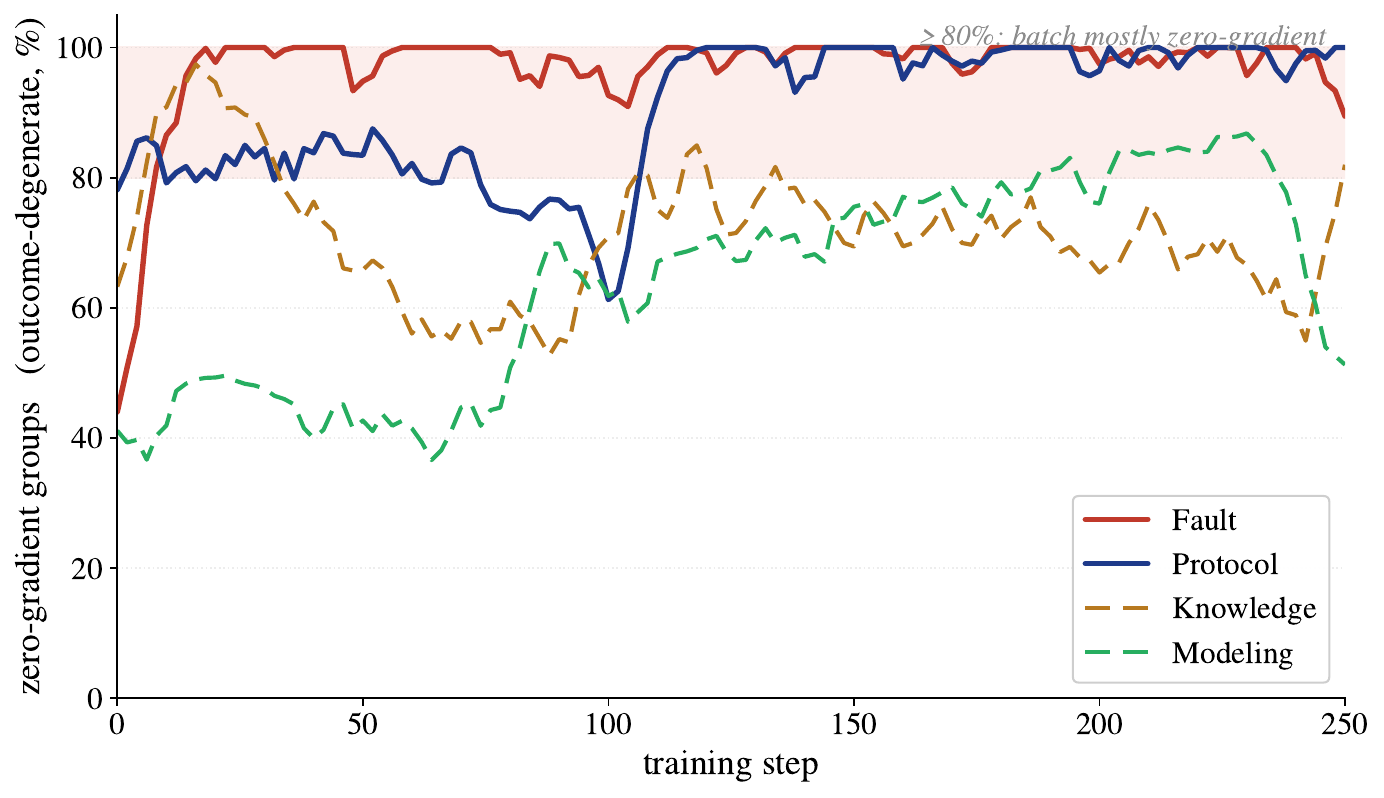}
    \caption{Zero std groups per axis during RL(GRPO) training with outcome reward.}
    \label{fig:zerostd}
\end{figure}
\begin{figure*}[t]
    \centering
    \includegraphics[width=1\linewidth]{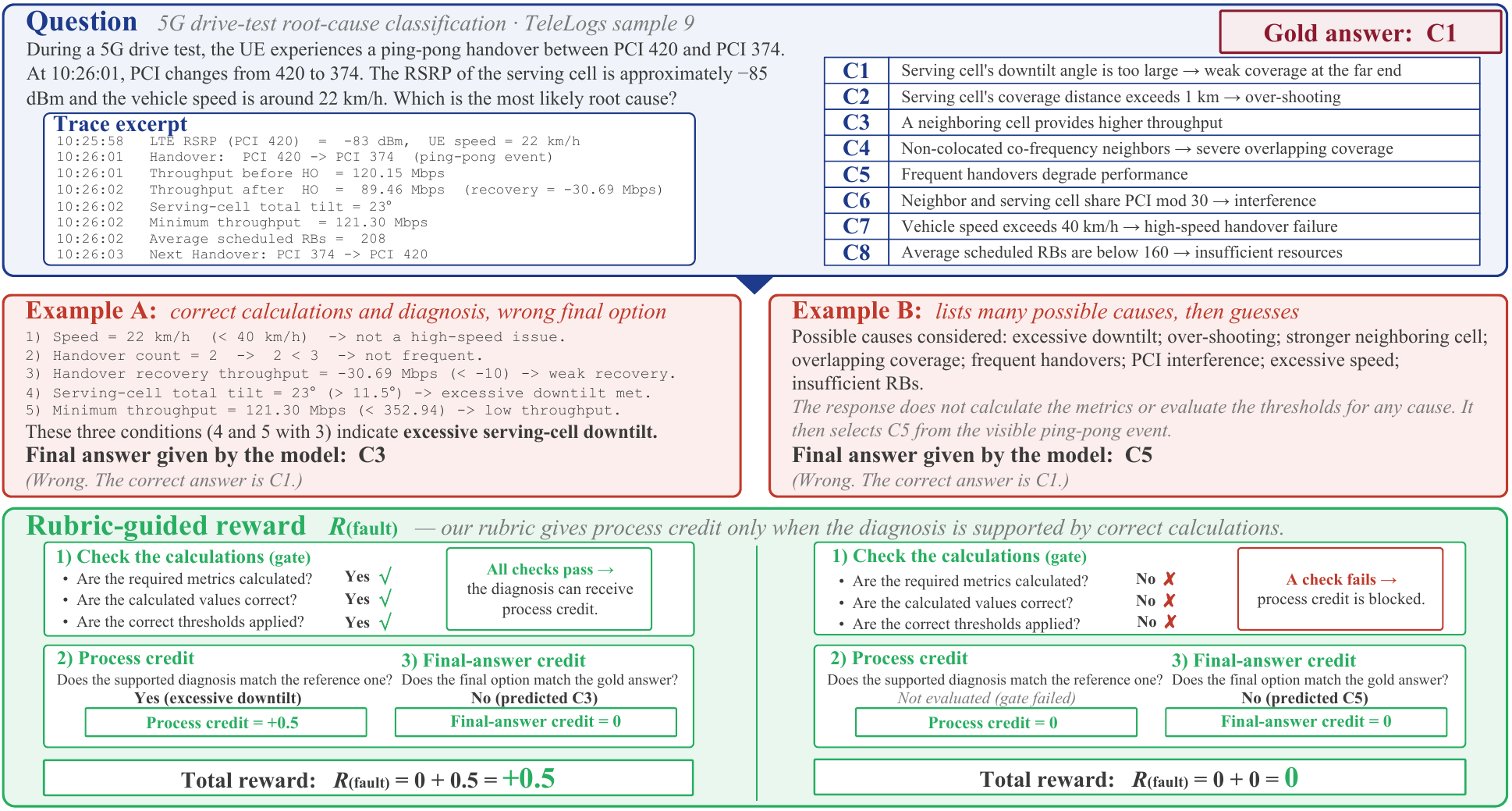}
    \caption{Fault axis case study on why rubric rewards are necessary.}
    \label{fig:rubric_case_study}
\end{figure*}

Reasoning across heterogeneous telecom tasks requires rewards that distinguish meaningful differences among sampled responses. Although terminal-answer rewards are simple and directly tied to task success, they provide little group-relative supervision when most responses to a prompt receive the same outcome. For example, two fault-diagnosis responses may share the same incorrect final option even though one contains verifier-supported calculations and the other merely guesses randomly.

\subsubsection{Outcome-only rewards are too sparse for hard telecom axes}
As stated in Sec~\ref{sec:analysis:coldstart}, group-relative RL relies on reward variation within each rollout group. If all rollouts for a prompt receive the same terminal reward, the reward standard deviation becomes zero, leaving no relative advantage signal.

Fig.~\ref{fig:zerostd} reports the percentage of rollout groups with zero outcome-reward variance under outcome-only GRPO training. Fault reasoning approaches near-complete outcome degeneracy and remains at a high level for most of training, and protocol axis also exhibits a high fraction of zero-variance groups in later stages. In contrast, knowledge and modeling retain more terminal-reward variation. Outcome-only rewards cannot distinguish useful reasoning behaviors when terminal outcomes are identical.

Under an outcome-only reward, dynamic sampling can remove zero-variance groups but cannot assign different credit to incorrect responses that contain different amounts of valid intermediate reasoning. We therefore equip protocol and fault tasks with rubric-guided process rewards that credit verifiable progress despite identical terminal outcomes.

\subsubsection{Grounded process rewards recover useful learning signals}
\begin{table}[t]
\centering
\caption{Reward-design ablation on 3GPP-TSG and TeleLogs with 9B backbone.}
\label{tab:reward_design_ablation}
\scriptsize
\setlength{\tabcolsep}{4pt}
\begin{tabular}{@{}>{\raggedright\arraybackslash}p{0.26\columnwidth}ccc>{\raggedright\arraybackslash}p{0.30\columnwidth}@{}}
\hline\hline
Reward design & 3GPP & TeleLogs & Mean & Failure mode \\
\hline
Outcome-only
& 75.0 & 58.0 & 66.5
& Low reward variance \\

Ungrounded process reward
& 76.0 & 65.2 & 70.6
& Rewards surface features; no verifier binding \\

Verifier-grounded rubric, ours
& \textbf{77.9} & \textbf{87.5} & \textbf{82.7}
& Verifier-grounded, calculation-gated \\
\hline
\end{tabular}
\end{table}

\begin{figure*}[!t]
    \centering
    \begin{subfigure}{0.44\textwidth}
        \centering
        \includegraphics[width=\linewidth]{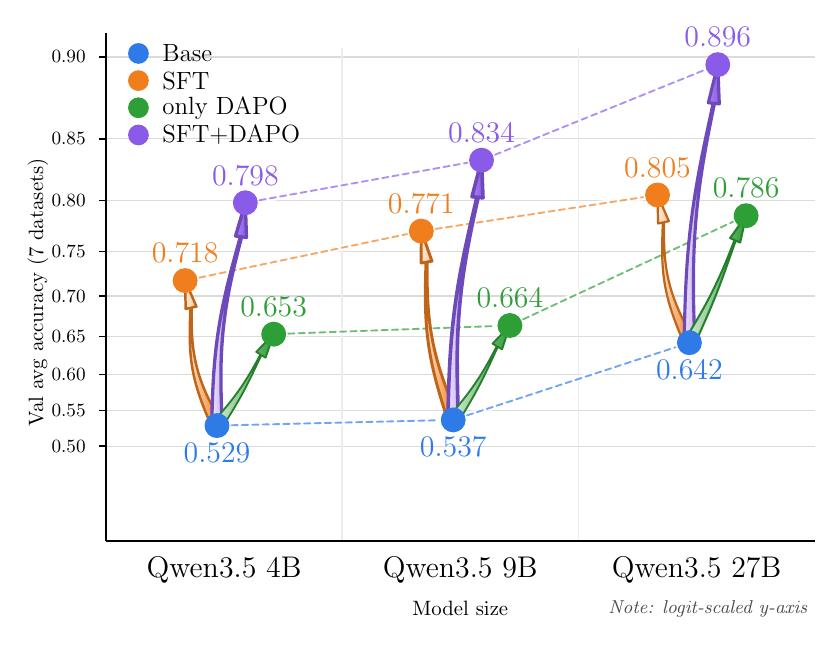}
        \caption{Model-size scaling.}
        \label{fig:model_size_scaling}
    \end{subfigure}\hfill
    \begin{subfigure}{0.44\textwidth}
        \centering
        \includegraphics[width=\linewidth]{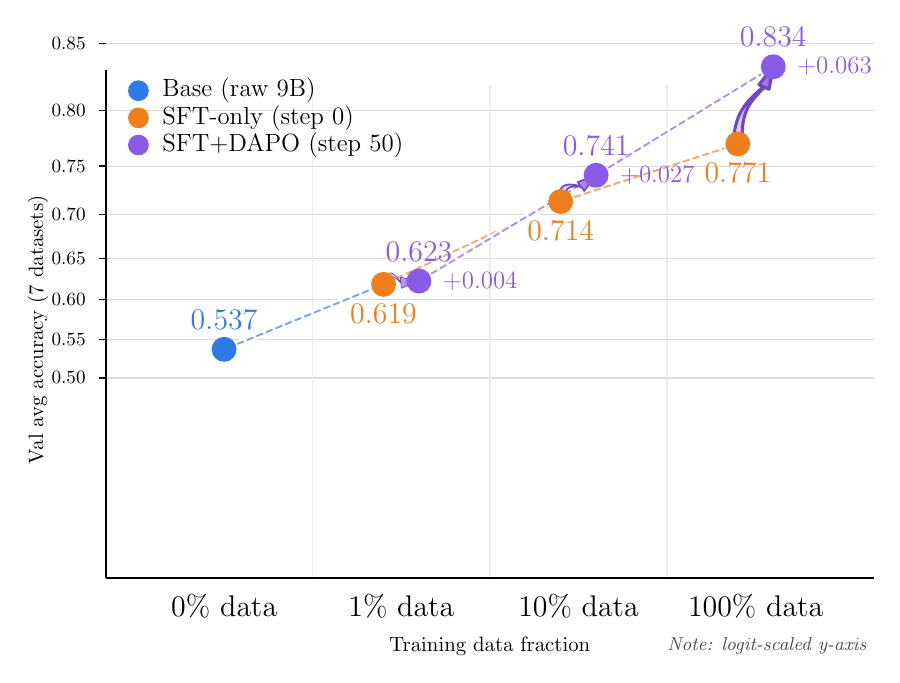}
        \caption{Data-size scaling.}
        \label{fig:data_size_scaling}
    \end{subfigure}
    \caption{Scaling behavior of \textit{TelecomGPT-R1-9B}.}
    \label{fig:scaling}
\end{figure*}

Process credit is useful only when the credited reasoning is verifiable. 
A ungrounded process reward may assign credit for mentioning relevant telecom concepts, listing many possible causes, or producing calculations that are unrelated to the selected diagnosis. Our rewards instead bind process credit to task-specific evidence and deterministic checks.

For fault diagnosis, the calculations form a binary gate for diagnosis-level credit, as defined in Eqs.~\eqref{eq:reward_fault} and~\eqref{eq:fault_calculation_check}. The gate passes only when the response reports all required metrics, reproduces their replayed values, and applies the correct thresholds. Process credit is then awarded only if the supported diagnosis matches the reference.

Fig.~\ref{fig:rubric_case_study} compares two responses with incorrect final options. Example~A correctly computes the required metrics, applies their thresholds, and diagnoses excessive serving-cell downtilt, but outputs C3 instead of the gold option C1. It therefore receives no outcome reward but earns $0.5$ process credit. Example~B lists several plausible causes and selects C5 from a visible ping-pong event without verifying the required metrics or diagnostic conditions. Its calculation gate fails, so it receives no process credit. 
The comparison illustrates the complementary roles of the two reward components. Outcome verification enforces final-answer correctness, while calculation-gated process credit distinguishes verifier-supported intermediate reasoning from unsupported guesses.

The protocol rubric follows the same principle. It gives limited credit when the gold WG appears among the candidates or the prediction belongs to the correct family, while penalizing responses that initially identify the gold group but later reject it without valid evidence. Thus, protocol credit reflects both candidate identification and the consistency of the final decision, rather than merely rewarding the presence of relevant WG names.

Table~\ref{tab:reward_design_ablation} compares outcome-only verification, ungrounded process rewards, and our verifier-grounded rubrics, with all accuracies measured at the same training step. On 3GPP, accuracy increases from $75.0\%$ with outcome-only rewards to $76.0\%$ with ungrounded process rewards and $77.9\%$ with our grounded rubric. On TeleLogs, the corresponding scores are $58.0\%$, $65.2\%$, and $87.5\%$. The two-benchmark mean consequently rises from $66.5\%$ to $70.6\%$ and then $82.7\%$. 


These results show that hard telecom axes require dense process credit, but that credit must be bound to axis-specific verifiers. Outcome-only rewards leave many rollout groups with zero reward variance and therefore provide no group-relative signal for those prompts. The ungrounded process baseline provides denser feedback but yields smaller gains, consistent with its ability to reward surface features that are not necessarily tied to a correct diagnosis or WG decision. 


\subsection{On Scaling Telecom Reasoners}
\label{sec:analysis:scaling}

Finally, we study how telecom reasoning performance scales with model capacity and data size. We observe that clear scaling trends emerge along both dimensions. 

\subsubsection{Scaling model size}

Fig.~\ref{fig:model_size_scaling} compares raw, SFT-only, DAPO-only, and SFT+DAPO models at the 4B, 9B, and 27B scales. The complete recipe improves average accuracy from $52.9\%$ to $79.8\%$, from $53.7\%$ to $83.4\%$, and from $64.2\%$ to $89.6\%$, respectively. Moreover, the post-trained 4B model exceeds the raw 27B backbone by $15.6$ pp, showing that model scaling raises the performance ceiling but cannot replace telecom-specific post-training.

As shown in Table~\ref{tab:main}, the gains of SFT and RL are axis-dependent. Across the three scales, SFT improves protocol reasoning by $44.7$, $42.0$, and $26.7$ pp and modeling by $50.0$, $56.6$, and $47.6$ pp, respectively. In contrast, DAPO provides its largest gains over SFT on fault reasoning, reaching $36.4$, $32.3$, and $34.2$ pp. Thus, SFT primarily installs domain knowledge and derivation procedures, while RL strengthens the integration of metrics, rules, and diagnostic decisions.

Additional capacity benefits the most complex tasks. From the final 4B to 27B models, accuracy increases by $13.5$ pp on fault and by $12.2$ pp on average across modeling, compared with $9.8$ pp on knowledge and $7.2$ pp across protocol. Larger models therefore contribute most to multi-step computation and rule-intensive reasoning.

\subsubsection{Scaling data size}

We next scale the SFT and RL data jointly while fixing the 9B backbone. The $1\%$, $10\%$, and full settings contain $924/125$, approximately $9{,}238/1{,}250$, and $92{,}380/12{,}500$ SFT examples/RL prompts, respectively.

As shown in Fig.~\ref{fig:data_size_scaling}, SFT and SFT+DAPO improve from $61.8\%/62.2\%$ at $1\%$ data to $71.4\%/74.1\%$ at $10\%$ and $77.1\%/83.4\%$ with the full dataset. Even the $1\%$ setting outperforms the raw 9B backbone at $53.7\%$, while moving from $10\%$ to the full dataset adds another $9.3$ pp, indicating that data scaling has not yet saturated.


The gain from DAPO over SFT also grows with data size, increasing from $0.4$ pp at $1\%$ to $2.7$ pp at $10\%$ and $6.3$ pp with the full dataset. Thus, SFT dominates under limited data, whereas broader coverage creates more opportunities for RL improvement. Overall, model capacity raises the reasoning ceiling, while data scale expands task coverage and amplifies the benefit of RL.

\section{Conclusion}\label{sec:conclusion}

We presented \textit{TelecomGPT-R1}, an open source family of telecom reasoners trained across protocol, knowledge, modeling, and fault tasks. Its unified post-training framework combines axis-aware data construction, an SFT cold start, and DAPO with task-routed, verifier-grounded rubric rewards. These components enable one policy to reason across heterogeneous telecom evidence and achieve state-of-the-art performance on the GSMA Open Telco Leaderboard. Our analysis yields four main takeaways. Effective RL requires a knowledge-grounded cold start, while broad competence depends on source-diverse training. DAPO further improves multi-source optimization by focusing updates on prompts that retain useful reward variation. For difficult axes, dense process credit must be grounded in axis-specific verifiers to recognize valid intermediate progress without reinforcing unsupported reasoning. Finally, model capacity and data coverage are complementary, where the former raises the reasoning ceiling, and the latter broadens task coverage and increases the benefit of RL. These findings highlight the need to coordinate data construction, reasoning supervision, verification, and policy optimization. Future work will extend \textit{TelecomGPT-R1} to broader tasks and evidence sources, develop it as the reasoning core of verifiable telecom agents, and incorporate multimodal evidence toward more comprehensive telecom intelligence.

\bibliographystyle{IEEEtran}
\bibliography{bib}

\end{document}